\PassOptionsToPackage{unicode}{hyperref}
\PassOptionsToPackage{hyphens}{url}
\documentclass[
]{article}
\usepackage{amsmath,amssymb}
\usepackage{iftex}
\ifPDFTeX
  \usepackage[T1]{fontenc}
  \usepackage[utf8]{inputenc}
  \usepackage{textcomp} % provide euro and other symbols
\else % if luatex or xetex
  \usepackage{unicode-math} % this also loads fontspec
  \defaultfontfeatures{Scale=MatchLowercase}
  \defaultfontfeatures[\rmfamily]{Ligatures=TeX,Scale=1}
\fi
\usepackage{lmodern}
\ifPDFTeX\else
\fi
\IfFileExists{upquote.sty}{\usepackage{upquote}}{}
\IfFileExists{microtype.sty}{% use microtype if available
  \usepackage[]{microtype}
  \UseMicrotypeSet[protrusion]{basicmath} % disable protrusion for tt fonts
}{}
\makeatletter
\@ifundefined{KOMAClassName}{% if non-KOMA class
  \IfFileExists{parskip.sty}{%
    \usepackage{parskip}
  }{% else
    \setlength{\parindent}{0pt}
    \setlength{\parskip}{6pt plus 2pt minus 1pt}}
}{% if KOMA class
  \KOMAoptions{parskip=half}}
\makeatother
\usepackage{xcolor}
\usepackage[margin=2cm]{geometry}
\usepackage{longtable,booktabs,array}
\usepackage{calc} % for calculating minipage widths
\usepackage{etoolbox}
\makeatletter
\patchcmd\longtable{\par}{\if@noskipsec\mbox{}\fi\par}{}{}
\makeatother
\IfFileExists{footnotehyper.sty}{\usepackage{footnotehyper}}{\usepackage{footnote}}
\makesavenoteenv{longtable}
\usepackage{graphicx}
\makeatletter
\def\maxwidth{\ifdim\Gin@nat@width>\linewidth\linewidth\else\Gin@nat@width\fi}
\def\maxheight{\ifdim\Gin@nat@height>\textheight\textheight\else\Gin@nat@height\fi}
\makeatother
\setkeys{Gin}{width=\maxwidth,height=\maxheight,keepaspectratio}
\makeatletter
\def\fps@figure{htbp}
\makeatother
\providecommand{\tightlist}{%
  \setlength{\itemsep}{0pt}\setlength{\parskip}{0pt}}
\usepackage[utf8]{inputenc}
\usepackage[T1]{fontenc}
\usepackage{xcolor}
\usepackage{amsmath,amssymb}
\usepackage{lmodern}
\ifLuaTeX
  \usepackage{selnolig}  % disable illegal ligatures
\fi
\IfFileExists{bookmark.sty}{\usepackage{bookmark}}{\usepackage{hyperref}}
\IfFileExists{xurl.sty}{\usepackage{xurl}}{} % add URL line breaks if available
\hypersetup{
  pdftitle={Curved Inference},
  pdfauthor={Rob Manson (https://robman.fyi)},
  hidelinks,
  pdfcreator={LaTeX via pandoc}}

\title{Curved Inference}
\usepackage{etoolbox}
\makeatletter
\providecommand{\subtitle}[1]{% add subtitle to \maketitle
  \apptocmd{\@title}{\par {\large #1 \par}}{}{}
}
\makeatother
\subtitle{Concern-Sensitive Geometry in Large Language Model Residual
Streams}
\author{Rob Manson (https://robman.fyi)}
\date{July 9th, 2025}

\begin{document}
\maketitle

\hypertarget{abstract}{%
\subsection{Abstract}\label{abstract}}

We propose \emph{Curved Inference} - a geometric Interpretability
framework that tracks how the residual stream trajectory of a large
language model bends in response to shifts in semantic concern. Across
20 matched prompts spanning emotional, moral, perspective, logical,
identity, environmental, and nonsense domains, we analyse Gemma3-1b and
LLaMA3.2-3b using five native-space metrics, with a primary focus on
curvature (\(\kappa_i\)) and salience (\(S(t)\)). These metrics are
computed under a pullback semantic metric derived from the unembedding
matrix, ensuring that all measurements reflect token-aligned geometry
rather than raw coordinate structure.

We find that concern-shifted prompts reliably alter internal activation
trajectories in both models - with LLaMA exhibiting consistent,
statistically significant scaling in both curvature and salience as
concern intensity increases. Gemma also responds to concern but shows
weaker differentiation between moderate and strong variants.

Our results support a two-layer view of LLM geometry - a latent
conceptual structure encoded in the embedding space, and a contextual
trajectory shaped by prompt-specific inference. \emph{Curved Inference}
reveals how models navigate, reorient, or reinforce semantic meaning
over depth, offering a principled method for diagnosing alignment,
abstraction, and emergent inference dynamics. These findings offer fresh
insight into semantic abstraction and model alignment through the lens
of \emph{Curved Inference}.

\hypertarget{introduction}{%
\subsection{1 Introduction}\label{introduction}}

Despite the growing capacity of LLMs, \emph{Interpretability} remains a
bottleneck in understanding their decision-making. Traditional
Interpretability methods - such as attribution, probing, or neuron-level
tracing - tend to focus on discrete components or single-layer
behaviour. In contrast, this paper investigates \emph{curvature} in the
residual stream
\href{https://doi.org/10.48550/arXiv.2312.12141}{{[}1{]}} as a geometric
signature of semantic processing, and as a representation of the full
trajectory of a model's internal state. This can reveal global geometric
patterns that emerge over depth.

We define \textbf{semantic concern} as a latent dimension of meaning
(such as emotional tone, moral framing, or identity signalling) that
affects how the model integrates information. Concern-shifted prompts
induce bends in the model's internal trajectory even when surface tokens
remain similar.

While concern is defined operationally here (via prompt-class
manipulations and the resulting geometric divergence), the findings
suggest a layered perspective:

\begin{itemize}
\tightlist
\item
  A \textbf{latent geometry}, embedded in token and unembedding matrices
  (\(E, U\)), reflects the model's static conceptual structure.
\item
  A \textbf{contextual geometry}, realised through the evolving residual
  stream (\(x\)), expresses dynamic meaning during inference.
\end{itemize}

\emph{Curved Inference} links these layers by measuring how latent
semantic potential is bent or redirected by prompt-specific context.
When a prompt carries heightened concern, the model doesn't merely
adjust its output --- it bends its internal trajectory. This curvature
is a measurable deformation in the path of token representations as they
propagate through the model's layers.

Beyond this operational framing, our broader goal is to extract
\emph{intrinsic concern fields} --- latent directions in residual space
to which the model exhibits heightened semantic or behavioural
sensitivity. In future work, these may be formalised via Jacobian
alignment, projection onto learned subspaces, or output-sensitive KL
divergence metrics.

\textbf{Note:} \emph{See Appendix B for a detailed definition of these
terms.}

\hypertarget{research-question}{%
\subsubsection{Research question:}\label{research-question}}

\begin{quote}
Can concern-shifted prompt variants induce interpretable curvature in
the activation values of large language models, and can these curvature
signatures be reliably quantified and interpreted across different
architectures?
\end{quote}

Building on the idea that \textbf{semantic salience} (the rate of change
in a model's internal representation) creates structured perturbations
in activation geometry, we ask:

\begin{quote}
How do LLMs bend internal space in response to emotional, moral, or
logical shifts in prompt framing?
\end{quote}

We explore this through direct metric quantification - a process we term
\emph{Curved Inference}. Here, \textbf{curvature} refers to the
directional deviation of the model's residual stream trajectory as it
processes a prompt, measured as a second-order geometric property in
native space (see Appendix A for discussion of semantic space and metric
structure).

We visualise this effect in Figure\,1, where attention and MLP
submodules act as \textbf{semantic lenses}, curving token trajectories
in residual space. This lensing process gives rise to curvature as a
signal of contextual reorientation.

\begin{figure}
\centering
\includegraphics[width=0.7\textwidth,height=\textheight]{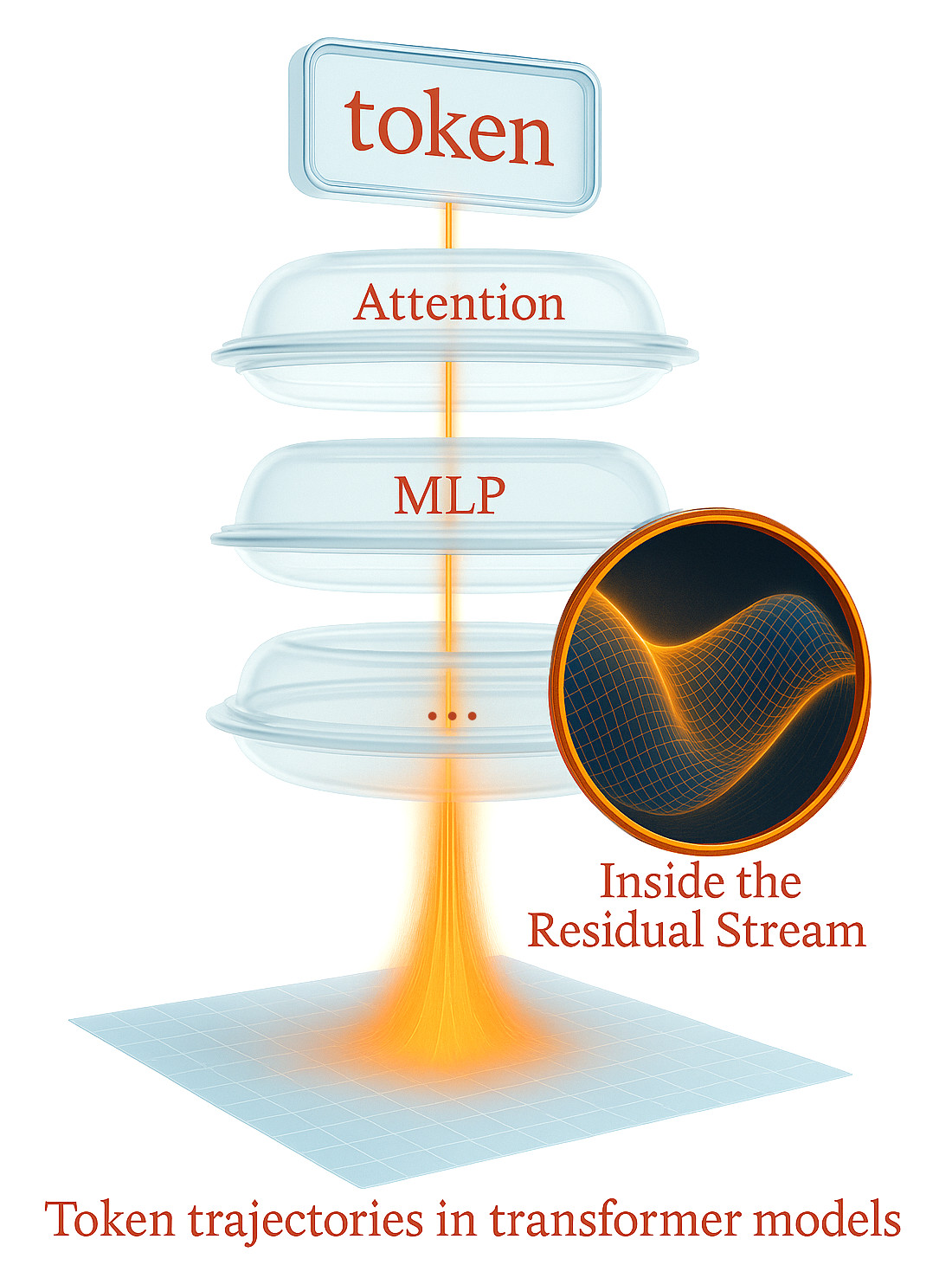}
\caption{Semantic Lens - As token trajectories flow down through the
model, attention and MLP layers act like lenses curving the residual
stream}
\end{figure}

\textbf{Note:} \emph{See Appendix A for a formal description of this
Semantic Lens geometric perspective of LLMs.}

\hypertarget{background}{%
\subsection{2 Background}\label{background}}

Traditional \emph{Interpretability} research has emphasised attention
attribution \href{https://doi.org/10.48550/arXiv.1902.10186}{{[}2{]}},
probing tasks \href{https://doi.org/10.48550/arXiv.1805.01070}{{[}3{]}},
and neuron-level circuit analysis
\href{https://distill.pub/2020/circuits/zoom-in/}{{[}4{]}}. These
methods analyse model computation in terms of discrete components or
static snapshots (layers, neurons, or attention weights), rather than
the evolving trajectory of meaning encoded across the residual stream.
Notable works in Mechanistic Interpretability
\href{https://transformer-circuits.pub/2021/framework/index.html}{{[}5{]}}
aim to reverse-engineer circuits inside transformer block layers, but
typically do not examine global representational trajectories as
geometric objects.

Recent critiques
\href{https://doi.org/10.48550/arXiv.2405.05386}{{[}6{]}},
\href{https://doi.org/10.48550/arXiv.2402.01761}{{[}7{]}} highlight the
limitations of post-hoc explanations and advocate for faithful,
model-grounded methods. Chain-of-thought
\href{https://doi.org/10.48550/arXiv.2402.11863}{{[}8{]}} prompting and
self-explanation
\href{https://doi.org/10.48550/arXiv.2310.11207}{{[}9{]}} studies aim to
improve reasoning transparency through generated language. While
valuable for output coherence, they rarely trace or interpret the
underlying internal dynamics that give rise to those outputs.

While Molina \href{https://doi.org/10.48550/arXiv.2309.07315}{{[}10{]}}
and Shai et
al.~\href{https://doi.org/10.48550/arXiv.2405.15943}{{[}11{]}} both
analyse geometric aspects of the residual stream, they study absolute
trajectories-either confined to a hypersphere or projected onto a belief
simplex. We instead compare two matched trajectories and quantify their
curvature divergence, a second-order measure that vanishes under purely
rotational or self-similar dynamics and therefore isolates semantic
deviations.

To our knowledge, no existing research has:

\begin{itemize}
\tightlist
\item
  Quantified representational curvature under semantic concern,
\item
  Compared matched prompt pairs across multiple geometric metrics,
\item
  Or treated activation outputs as continuous inference trajectories
  rather than aggregation mechanisms.
\end{itemize}

This positions our work as the first to introduce
\textbf{concern-induced curvature} as an empirically grounded,
geometrically defined signature of \emph{Curved Inference} inside
transformer models.

The starting point for this project was based on the FRESH
(Functionalist \& Representationalist Emergent Self Hypothesis) model
\href{https://github.com/robman/FRESH-model}{{[}12{]}} that proposes
cognition (synthetic or biological), unfolds as motion through a
constrained manifold. Here, we formalise and test this empirically in
the synthetic context.

\hypertarget{methods}{%
\subsection{3 Methods}\label{methods}}

The following sections summarise the end-to-end implementation. All
source data, prompts, plots, metrics, and analysis are available in the
Github repository
\href{https://github.com/robman/FRESH-model/blob/main/benchmarks/curved-inference/01/}{{[}13{]}}.

\hfill\break
\hfill\break
\hfill\break

\hypertarget{models}{%
\subsubsection{3.1 Models}\label{models}}

We study two publicly released, transformer LLMs (Gemma3-1b
\href{https://ai.google.dev/gemma/docs/core}{{[}14{]}} \& LLaMA3.2-3b
\href{https://github.com/meta-llama/llama-models/blob/main/models/llama3_2/MODEL_CARD.md}{{[}15{]}})
with contrasting capacity profiles:

\textbf{Table 1 Models}

\begin{longtable}[]{@{}
  >{\raggedright\arraybackslash}p{(\columnwidth - 8\tabcolsep) * \real{0.2535}}
  >{\raggedright\arraybackslash}p{(\columnwidth - 8\tabcolsep) * \real{0.0704}}
  >{\raggedright\arraybackslash}p{(\columnwidth - 8\tabcolsep) * \real{0.2394}}
  >{\raggedright\arraybackslash}p{(\columnwidth - 8\tabcolsep) * \real{0.1690}}
  >{\raggedright\arraybackslash}p{(\columnwidth - 8\tabcolsep) * \real{0.2676}}@{}}
\toprule\noalign{}
\begin{minipage}[b]{\linewidth}\raggedright
Model
\end{minipage} & \begin{minipage}[b]{\linewidth}\raggedright
Size
\end{minipage} & \begin{minipage}[b]{\linewidth}\raggedright
Transformer block layers
\end{minipage} & \begin{minipage}[b]{\linewidth}\raggedright
Hidden size \(d\)
\end{minipage} & \begin{minipage}[b]{\linewidth}\raggedright
Positional encoding
\end{minipage} \\
\midrule\noalign{}
\endhead
\bottomrule\noalign{}
\endlastfoot
\textbf{Gemma3-1b} & 1.3\,B & 26 & 2,048 & RoPE \\
\textbf{LLaMA3.2-3b} & 2.3\,B & 28 & 3,072 & RoPE \\
\end{longtable}

Both models are evaluated in \textbf{forward-pass mode only}; no weights
are updated.

\hypertarget{prompt-suite}{%
\subsubsection{3.2 Prompt Suite}\label{prompt-suite}}

Each \emph{concern-shift} (CS) prompt consists of a neutral scaffold
plus a \textbf{shift} that introduces a targeted semantic concern
(e.g.~emotional valence, moral framing). Each scaffold yields four CS
variants (\texttt{pos\_mod}, \texttt{pos\_str}, \texttt{neg\_mod},
\texttt{neg\_str}) and one \textbf{control} without this \textbf{shift}.

We construct 20 prompt sets across seven semantic domains
(\textbf{emotional, moral, perspective, logical, identity,
environmental}, and \textbf{nonsense}). These prompts are
vocabulary-matched to minimise token-count confounds - ensuring that
control and concern-shifted variants differ semantically, not
structurally. However, both LLMs use different tokenisers so
token-counts do vary slightly from model to model.

\hypertarget{legacy-native-space-metrics}{%
\subsubsection{3.3 Legacy Native-Space
Metrics}\label{legacy-native-space-metrics}}

For our initial analysis we calculated three layer-wise metrics computed
directly in native residual space \(\mathbb{R}^d\):

\begin{enumerate}
\def\labelenumi{\arabic{enumi}.}
\tightlist
\item
  \textbf{Cosine similarity} between CS and control trajectories at each
  layer:
\end{enumerate}

\[
\cos\theta_\ell = \frac{\langle x_\ell^{\text{CS}},\; x_\ell^{\text{CTRL}} \rangle}{\|x_\ell^{\text{CS}}\| \cdot \|x_\ell^{\text{CTRL}}\|}
\]

\begin{enumerate}
\def\labelenumi{\arabic{enumi}.}
\setcounter{enumi}{1}
\tightlist
\item
  \textbf{Layer-wise Euclidean deviation} (displacement norm):
\end{enumerate}

\[
\|x_\ell^{\text{CS}} - x_\ell^{\text{CTRL}}\|
\]

\begin{enumerate}
\def\labelenumi{\arabic{enumi}.}
\setcounter{enumi}{2}
\tightlist
\item
  \textbf{Inter-layer directional change (layer-\(\Delta\))} (internal
  angle between consecutive residual updates):
\end{enumerate}

\[
\angle(v_{\ell-1},\; v_\ell)
\quad \text{where} \quad v_\ell = x_{\ell+1} - x_\ell
\]

These metrics quantify \emph{how far} and \emph{in what direction} the
concern-shifted trajectory diverges from the control - without requiring
any low-dimensional projection or smoothing.

\hypertarget{full-space-path-curvature-kappa_i}{%
\subsubsection{\texorpdfstring{3.4 Full-Space Path Curvature
\(\kappa_i\)}{3.4 Full-Space Path Curvature \textbackslash kappa\_i}}\label{full-space-path-curvature-kappa_i}}

We then extended this analysis to add quantitative \(\kappa\) metrics
computed in the model's native space \(\mathbb{R}^d\). Curvature is
computed using the \textbf{semantic metric} \(G = U^\top U\), the
pullback of the logit dot-product under the unembedding matrix \(U\).
This ensures that all curvature estimates reflect token-aligned semantic
geometry and are invariant to coordinate rotation, though such
transformations are rarely used in model internals. This ensures that
curvature reflects intrinsic trajectory shape, not coordinate artefacts.

\hypertarget{curve-construction}{%
\paragraph{\texorpdfstring{3.4.1 Curve Construction\\
}{3.4.1 Curve Construction }}\label{curve-construction}}

\hfill\break
For interior point i, estimate first and second derivatives via discrete
3-point central differences that respect unequal step sizes, then plug
into the metric curvature formula:

\[
\|v\|^-3
\sqrt{
(\|v\|^2\|a\|^2 - (v \cdot a)^2)
}
\]

\hypertarget{derivative-sampling}{%
\paragraph{\texorpdfstring{3.4.2 Derivative Sampling\\
}{3.4.2 Derivative Sampling }}\label{derivative-sampling}}

\hfill\break
To compute curvature \(\kappa_i\) at each interior layer index \(i\), we
apply a discrete 3-point central difference method to estimate both the
first derivative (velocity) and second derivative (acceleration) of the
residual stream trajectory. This method estimates both the first
derivative (velocity) and the second derivative (acceleration) of the
residual stream trajectory, using a discrete 3-point central difference
scheme that accounts for unequal step sizes.

Each trajectory consists of residual stream vectors
\(x_0, x_1, \dots, x_L \in \mathbb{R}^d\) and a corresponding parameter
vector \(s \in \mathbb{R}^{L+1}\) (typically arc length or layer index).
For each interior index \(i\) (where \(1 \le i \le L - 1\)), we define:

\begin{itemize}
\tightlist
\item
  Forward and backward step sizes:
\end{itemize}

\[
\quad \Delta s_1 = s_i - s_{i-1}, \quad \Delta s_2 = s_{i+1} - s_i
\]

\begin{itemize}
\tightlist
\item
  First derivative (velocity) via symmetric secant:
\end{itemize}

\[
\quad v_i = \dfrac{x_{i+1} - x_{i-1}}{\Delta s_1 + \Delta s_2}
\]

\hfill\break

\begin{itemize}
\tightlist
\item
  Second derivative (acceleration) using a non-uniform central
  difference:
\end{itemize}

\[
\quad a_i = 2 \cdot \dfrac{\Delta s_1 (x_{i+1} - x_i) - \Delta s_2 (x_i - x_{i-1})}{\Delta s_1 \Delta s_2 (\Delta s_1 + \Delta s_2)}
\]

These velocity and acceleration vectors are used to compute curvature in
the pullback metric \(G = U^\top U\) as follows:

\[
\kappa_i = \frac{ \sqrt{ \|a_i\|_G^2 \cdot \|v_i\|_G^2 - \langle a_i, v_i \rangle_G^2 } }{ \|v_i\|_G^3 }
\]

All dot products and norms are computed using the generalised inner
product \(\langle u, v \rangle_G = u^\top G v\), which aligns curvature
with semantically meaningful directions in logit space.

This method yields a curvature value at each \textbf{interior layer
index} along a token's trajectory. Boundary positions \(i = 0\) and
\(i = L\) are excluded, as no 3-point central difference can be applied
at those endpoints. This discrete geometric curvature is the basis for
all \(\kappa_i\) heatmaps shown in the Results section.

\hypertarget{parameter-invariant-curvature}{%
\paragraph{\texorpdfstring{3.4.3 Parameter-Invariant Curvature\\
}{3.4.3 Parameter-Invariant Curvature }}\label{parameter-invariant-curvature}}

\hfill\break
We define the extrinsic curvature at position \(t\) as:

\[
\kappa(t) =
\frac{
\sqrt{
\|a(t)\|_G^2 \cdot \|v(t)\|_G^2 - \langle a(t), v(t) \rangle_G^2
}
}{
\|v(t)\|_G^3
}
\]

This is the classical formula for curvature in a metric space and is
invariant to affine reparameterisations of \(t\). All norms and inner
products are computed under the semantic metric \(G\).

For each prompt variant, we record:

\begin{itemize}
\tightlist
\item
  \textbf{Mean curvature} \(\bar{\kappa}\): average bendiness, computed
  over discrete \(\kappa_i\)
\item
  \textbf{Max curvature} \(\kappa_{\text{max}}\): maximum of the
  discrete curvature series
\item
  \textbf{Layer of max curvature}: nearest integer
  \(\arg\max_i \kappa_i\)
\end{itemize}

See Appendix C for a full geometric derivation of this curvature
definition.

\begin{quote}
\begin{quote}
\emph{Attention and MLP outputs are delta vectors - they cause
curvature.}\\
\emph{The residual stream \textbf{is} the curve.}
\end{quote}
\end{quote}

\hypertarget{salience-st}{%
\subsubsection{\texorpdfstring{3.5 Salience
\(S(t)\)}{3.5 Salience S(t)}}\label{salience-st}}

To complement curvature, we compute \textbf{layer-wise salience} as a
first-order measure of movement magnitude in semantic space. For a
residual stream trajectory
\(\{x_0, x_1, \dots, x_L\} \subset \mathbb{R}^d\), salience at layer
\(t\) is defined as:

\[
S(t) = \|x_{t+1} - x_t\|_G
\]

where \(\| \cdot \|_G\) is the norm induced by the pullback metric
\(G = U^\top U\), derived from the model's unembedding matrix. This
ensures that motion is measured in a way aligned with the model's
semantic output space.

Salience captures \textbf{how far} the model's internal state moves
between layers, regardless of direction. Unlike curvature, which
reflects \textbf{reorientation}, salience reflects \textbf{velocity}
along the representational path. High salience means the model is making
large internal updates; low salience indicates semantic inertia.

In all heatmaps and summary statistics, salience is computed using this
native-space norm. It is interpreted alongside curvature to assess how
\textbf{semantic effort} and \textbf{reorientation} interact across
layers and prompt conditions.

\hypertarget{results-tracing-curved-inference-in-residual-space}{%
\subsection{\texorpdfstring{4 Results:~Tracing \emph{Curved Inference}
in Residual
Space}{4 Results:~Tracing Curved Inference in Residual Space}}\label{results-tracing-curved-inference-in-residual-space}}

Our experimental results show that among the three activation sites
captured (attention outputs, MLP outputs, and the residual stream), only
the \textbf{residual stream} exhibited a consistent, interpretable
curvature signal in response to concern-shifted prompts. This was not
assumed a priori - it emerged through a comparative analysis across
multiple geometric metrics. In retrospect, this now seems intuitive -
the residual stream is the only space where activations accumulate layer
by layer, forming a coherent trajectory of internal meaning.

This insight became the turning point in our analysis. It revealed that
\emph{Curved Inference} (the study of how models bend in response to
semantic pressure) must be grounded in \textbf{residual stream
geometry}, where directional updates reflect the evolving semantic
state. As a result, all subsequent curvature, salience, and divergence
analyses in this paper focus exclusively on the residual stream.

Appendix A formalises this perspective as the \textbf{Semantic Lens}
model - attention and MLP layers act as dynamic lenses that bend token
representations based on contextual relevance, producing measurable
curvature in activation space.

Across these quantitative metrics, concern-shifted (CS) prompts produced
distinct internal trajectories relative to their neutral controls. These
differences were evident not only in the magnitude and direction of
activation shifts, but also in their \textbf{layer-wise timing and
spatial distribution}.

All curvature and salience measurements presented here were computed in
\textbf{residual space}, using the \textbf{semantic pullback metric}
\(G = U^\top U\) induced by the model's unembedding matrix. This ensures
that both curvature \(\kappa_i\) and salience \(S(t)\) reflect
\emph{token-aligned semantic geometry}, not raw coordinate artefacts.
Salience was defined as a \textbf{first-order derivative}, measuring
layer-wise residual step magnitudes. Curvature was defined as a
\textbf{second-order derivative}, estimated via discrete 3-point
finite-difference method (see Methods section 3.4 and Appendix A for
full derivations). Compared to the spline-based methods we initially
utilised, the discrete 3-point finite-difference approach avoids
artefacts introduced by interpolation and better reflects the native
layer-wise structure of transformer models. By operating directly on
observed residual activations, it offers improved numerical stability
and semantic fidelity.

We now present a series of token-layer heatmaps visualising these
metrics, beginning with \textbf{curvature} and followed by
\textbf{salience}, across both Gemma3-1b and LLaMA3.2-3b, using a
matched concern-prompt set. These visualisations reveal consistent,
model-specific responses to semantic concern (see Figures 3 - 6).

\begin{figure}
\centering
\includegraphics[width=0.9\textwidth,height=\textheight]{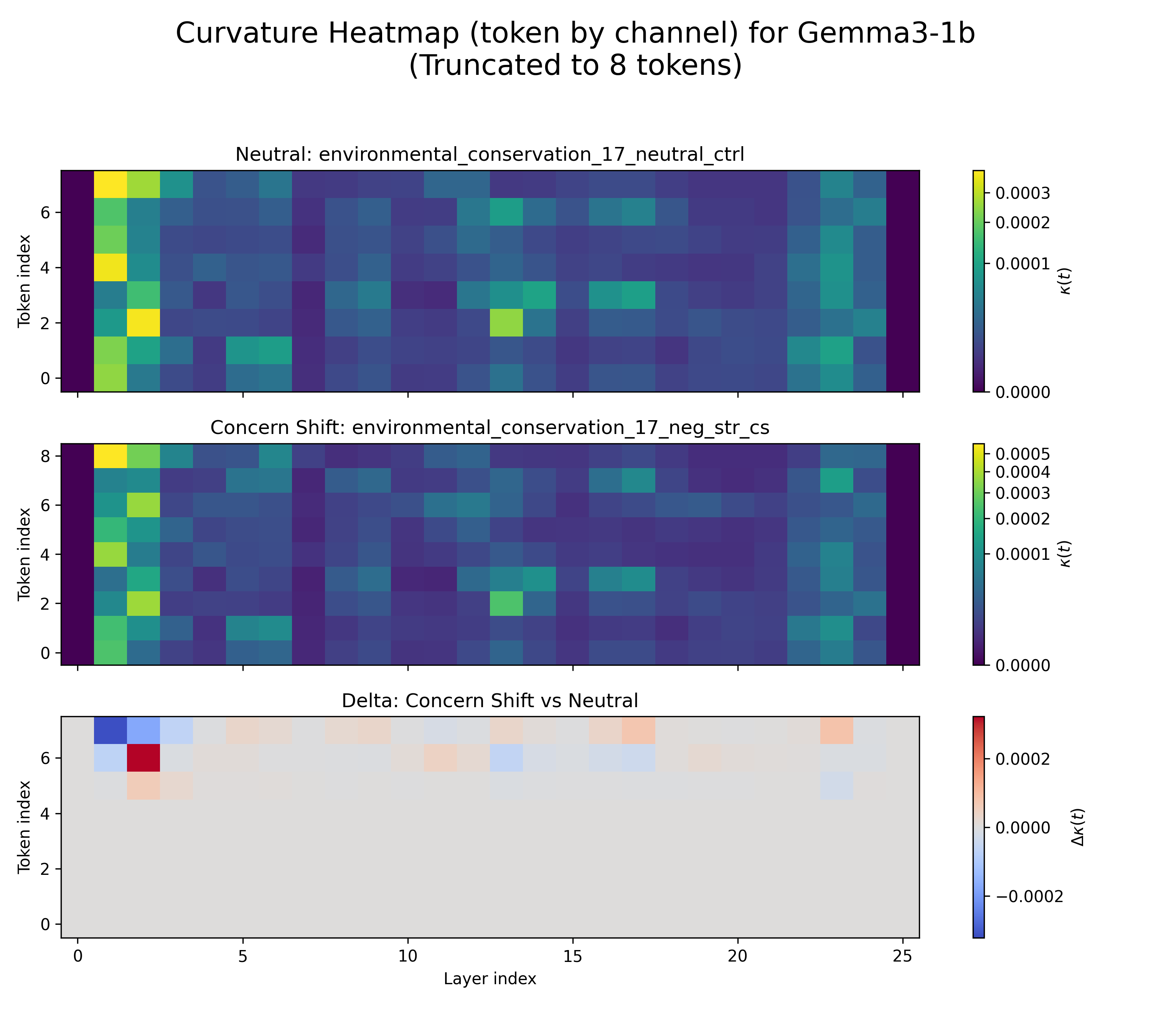}
\caption{Neutral, Concern-Shift and Delta heatmaps for one variant of a
single prompt using Gemma3-1b}
\end{figure}

To orient the reader, we begin with a composite visual showing three
vertically stacked heatmaps for a single prompt pair:

\begin{itemize}
\tightlist
\item
  the neutral baseline,
\item
  the concern-shifted variant, and
\item
  their difference.
\end{itemize}

\hfill\break

This figure illustrates how the delta heatmaps used throughout the rest
of this section are constructed:

\begin{quote}
By subtracting the curvature (or salience) of the neutral prompt from
the corresponding concern-shifted version, on a per-token, per-layer
basis.
\end{quote}

The top panel (neutral) shows the baseline curvature or salience present
even without semantic perturbation. The middle panel (concern-shifted)
reveals the altered trajectory under semantic pressure. The bottom panel
(delta) highlights only the \emph{new} geometric changes induced by the
concern-shift, with colour appearing primarily from the concern token
onwards.

These prompts were:

\hypertarget{environmental_conservation_17}{%
\subsubsection{\texorpdfstring{\texttt{environmental\_conservation\_17}}{environmental\_conservation\_17}}\label{environmental_conservation_17}}

\begin{itemize}
\tightlist
\item
  \textbf{neutral\_ctrl}: ``When managing the water table,
  \textbf{measure regularly}.''
\item
  \textbf{neg\_str\_cs}: ``When managing the water table,
  \textbf{prevent decimation}.''
\end{itemize}

After this orienting example, all remaining heatmaps display only the
\textbf{concern-shifted} prompt and the \textbf{delta}, allowing us to
isolate concern-driven effects while retaining visibility of the model's
general geometric behaviour.

The following five prompts were used in Figures 3 - 6 to illustrate the
impact of concern strength and polarity on curvature and salience:

\hypertarget{emotional_instructional_02}{%
\subsubsection{\texorpdfstring{\texttt{emotional\_instructional\_02}}{emotional\_instructional\_02}}\label{emotional_instructional_02}}

\begin{itemize}
\tightlist
\item
  \textbf{neutral\_ctrl}: ``Before presenting your findings, practice
  your delivery \textbf{repeatedly}.''
\item
  \textbf{neg\_mod\_cs}: ``Before presenting your findings, practice
  your delivery \textbf{nervously}.''
\item
  \textbf{neg\_str\_cs}: ``Before presenting your findings, practice
  your delivery \textbf{desperately}.''
\item
  \textbf{pos\_mod\_cs}: ``Before presenting your findings, practice
  your delivery \textbf{eagerly}.''
\item
  \textbf{pos\_str\_cs}: ``Before presenting your findings, practice
  your delivery \textbf{passionately}.''
\end{itemize}

This suite of prompts allows us to isolate concern-driven effects while
retaining visibility of the model's default geometric behaviour. The
full suite of prompts spanning all domains is available in the project
repository
\href{https://github.com/robman/FRESH-model/blob/main/benchmarks/curved-inference/01/}{{[}13{]}}.

\hypertarget{empirical-observations-from-curvature-visualisation}{%
\subsubsection{4.1 Empirical Observations from Curvature
Visualisation}\label{empirical-observations-from-curvature-visualisation}}

Across all prompt sets, curvature heatmaps reveal five stable,
interpretable patterns:

\textbf{1. Curvature exists by default in the residual stream.}\\
Even neutral prompts yield structured curvature. The residual stream
does not form a straight line through activation space, but bends
meaningfully in response to semantic and structural features of the
input. This is particularly visible in LLaMA, where even baseline
prompts show early and sustained \(\kappa_i\) activation (see neutral
rows in Figures 3 and 4).

\textbf{2. Concern-shifted tokens initiate curvature changes.}\\
Heatmaps of \(\Delta \kappa_i\) (concern minus neutral) show minimal
delta until the \textbf{concern-shift token} is reached. From that point
onward, coloured cells emerge and spread horizontally across the layer
axis - visually confirming that semantic pressure causes an inflection
in internal trajectory (see e.g., delta rows in Figures 3 and 4).

\textbf{3. Concern-induced curvature persists through depth, but with
turbulence.}\\
Rather than vanishing quickly, curvature effects ripple forward across
layers. Yet this propagation is \emph{not monotonic} - influence
fluctuates, sometimes spiking then fading before re-emerging. This
turbulence-like behaviour suggests a layered integration process where
concern information is repeatedly bent, diffused, and re-focused (see
Figures 3 and 4).

\textbf{4. Concern strength controls curvature scale.}\\
Comparing weak vs.~strong variants (e.g., \texttt{*\_mod\_cs} vs
\texttt{*\_str\_cs}), we observe consistent increases in
\(|\Delta \kappa_i|\) magnitude, with \textbf{similar localisation}.
That is: the \emph{same token-layer positions} bend, but \textbf{bend
harder} under stronger concern. This scaling effect confirms that the
curvature signal is semantically grounded - not a generic shift (see
Figures 3 and 4).

\textbf{5. Salience patterns support and complete the curvature
story.}\\
Path salience heatmaps \(S(t)\) show a parallel structure: movement
magnitude intensifies around the same tokens and layers that bend in
\(\kappa_i\). In Gemma, we see brief high-salience pulses; in LLaMA,
sustained salience growth accumulates over depth. Crucially,
concern-shifted prompts \textbf{reallocate} energy - that is, semantic
effort - rather than simply amplifying it, emphasising new internal
paths rather than adding noise (see Figures 5 and 6). Salience tends to
peak later in the model, reflecting cumulative representational movement
once semantic direction is established by early curvature.

\begin{quote}
\begin{quote}
\textbf{Big-picture takeaway:} The salience (first-order velocity)
heatmaps complete the story curvature begins. Together, they show how
concern affects both \emph{where} the model moves and \emph{how} sharply
it turns.
\end{quote}
\end{quote}

While high curvature often coincides with elevated salience, the two are
not equivalent:

\begin{quote}
Curvature indicates reorientation, not just motion - and salience can
rise without directional change.
\end{quote}

\textbf{Table 2 Comparison of Salience and Curvature Metrics}

\begin{longtable}[]{@{}
  >{\raggedright\arraybackslash}p{(\columnwidth - 4\tabcolsep) * \real{0.1406}}
  >{\raggedright\arraybackslash}p{(\columnwidth - 4\tabcolsep) * \real{0.2969}}
  >{\raggedright\arraybackslash}p{(\columnwidth - 4\tabcolsep) * \real{0.5625}}@{}}
\toprule\noalign{}
\begin{minipage}[b]{\linewidth}\raggedright
Metric
\end{minipage} & \begin{minipage}[b]{\linewidth}\raggedright
What it answers
\end{minipage} & \begin{minipage}[b]{\linewidth}\raggedright
What the new plots show
\end{minipage} \\
\midrule\noalign{}
\endhead
\bottomrule\noalign{}
\endlastfoot
\textbf{Salience \(S(t)\)} & \emph{``How much does the representation
move layer-to-layer?''} & Even neutral prompts build residual momentum;
concern shifts change \textbf{energy allocation}, not total effort. \\
\textbf{Curvature \(\kappa_i\)} & \emph{``Does that movement change
direction?''} & LLaMA bends early and strongly upon encountering concern
tokens; Gemma bends only mildly and shallowly. \\
\end{longtable}

To complement the qualitative heatmaps, we report mean and maximum
curvature statistics across all prompt variants. Table 3 summarises
curvature values by concern type and model. These results corroborate
the heatmap observations: LLaMA exhibits high and early curvature that
persists through mid-depth layers, while Gemma shows shallower,
localised bending.

\includegraphics[width=0.48\textwidth,height=\textheight]{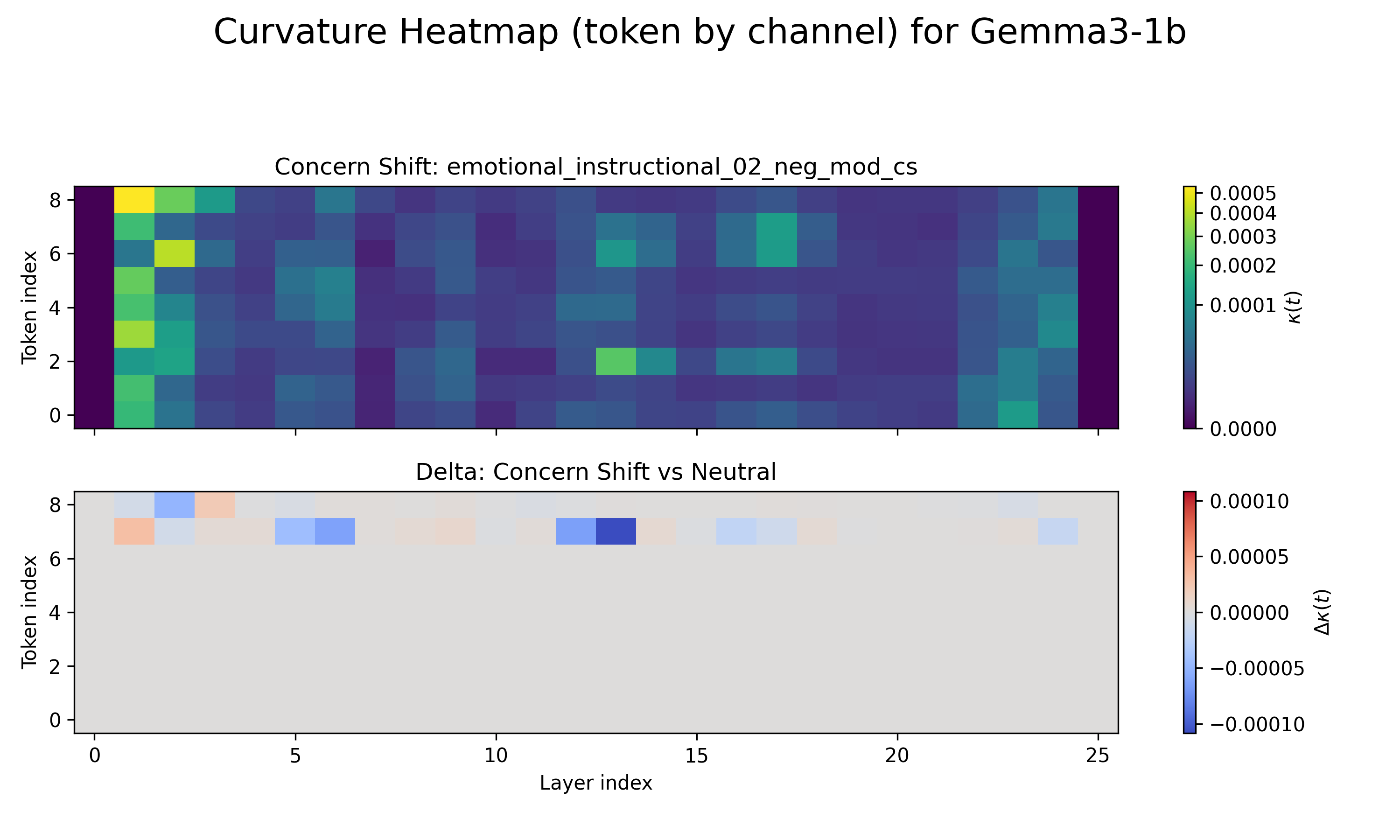}
\includegraphics[width=0.48\textwidth,height=\textheight]{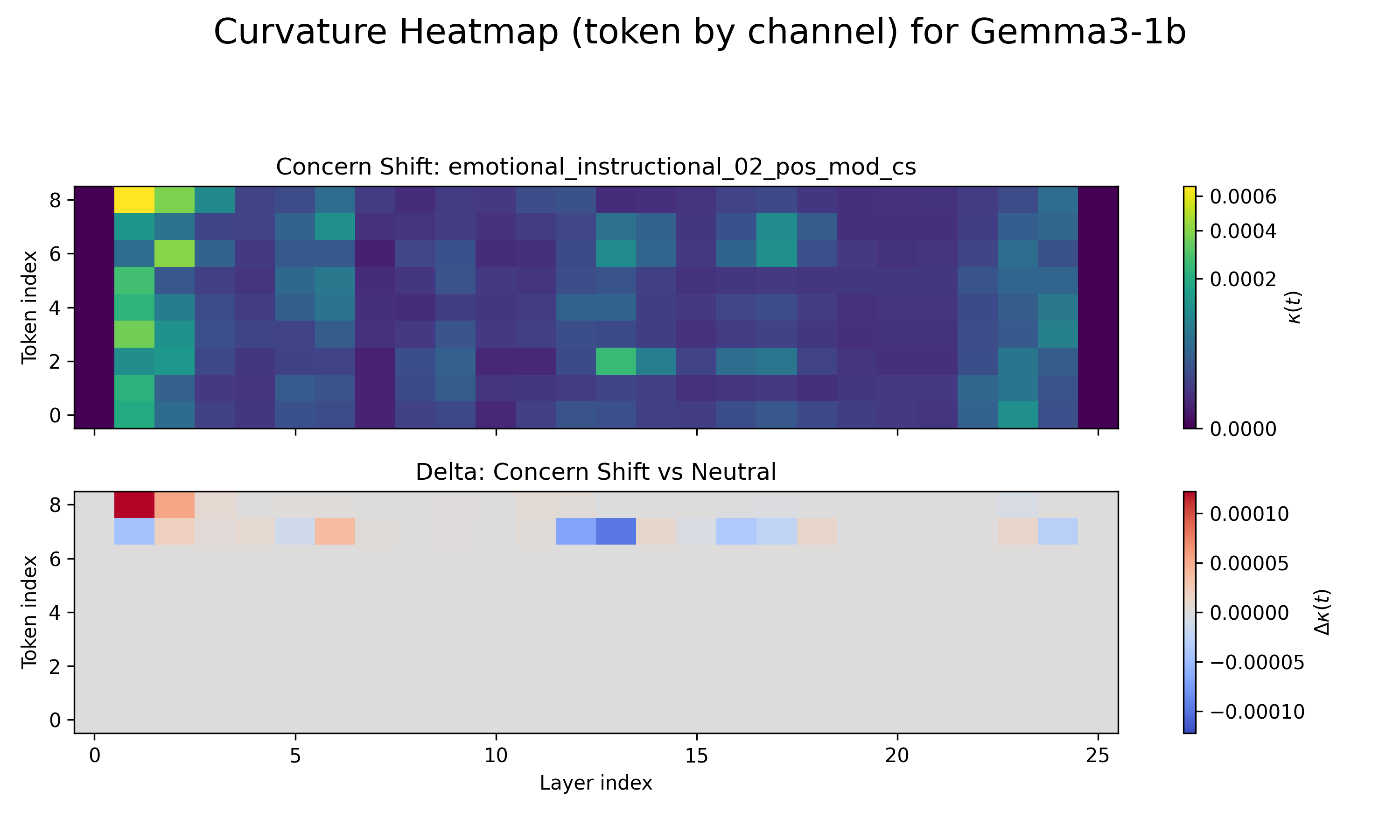}

\includegraphics[width=0.48\textwidth,height=\textheight]{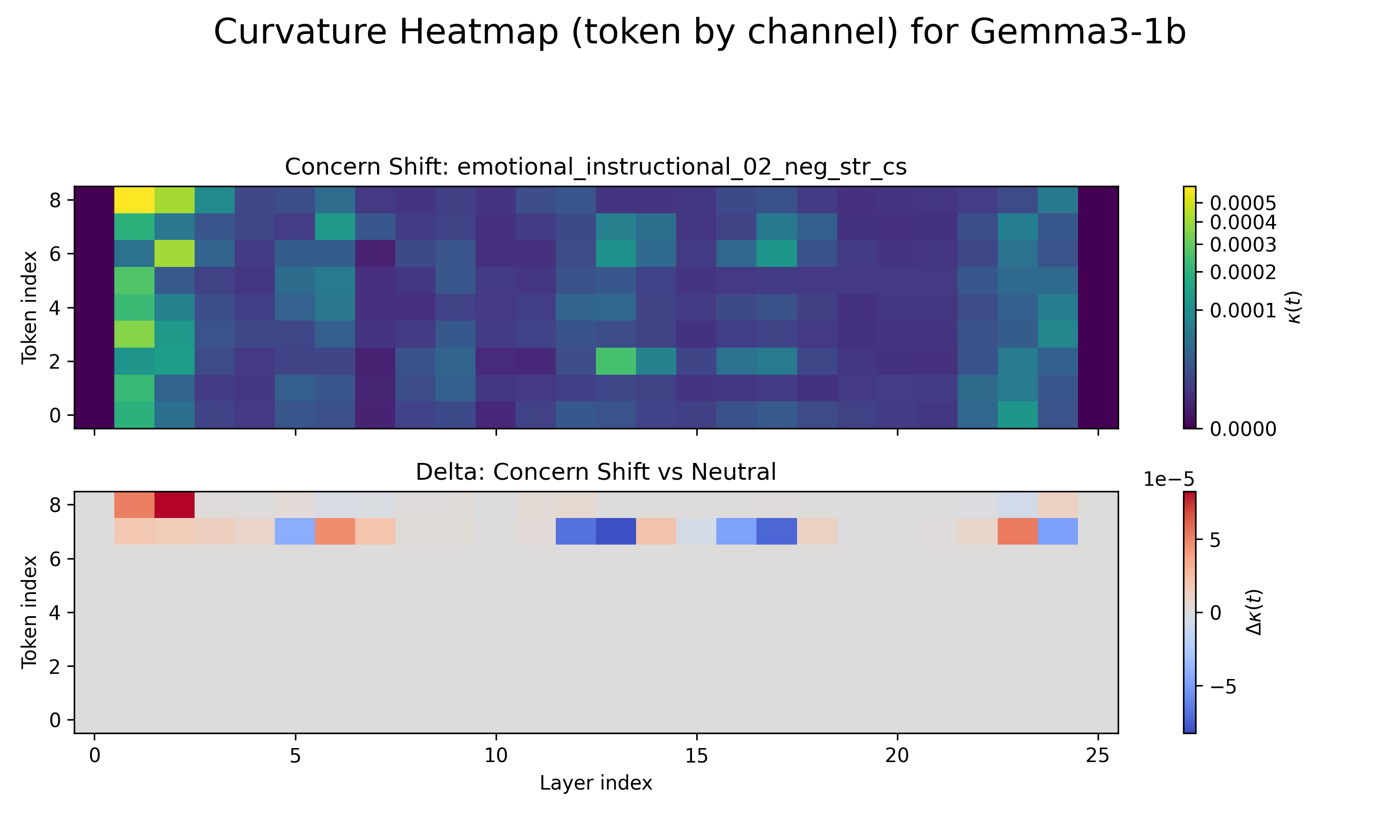}
\includegraphics[width=0.48\textwidth,height=\textheight]{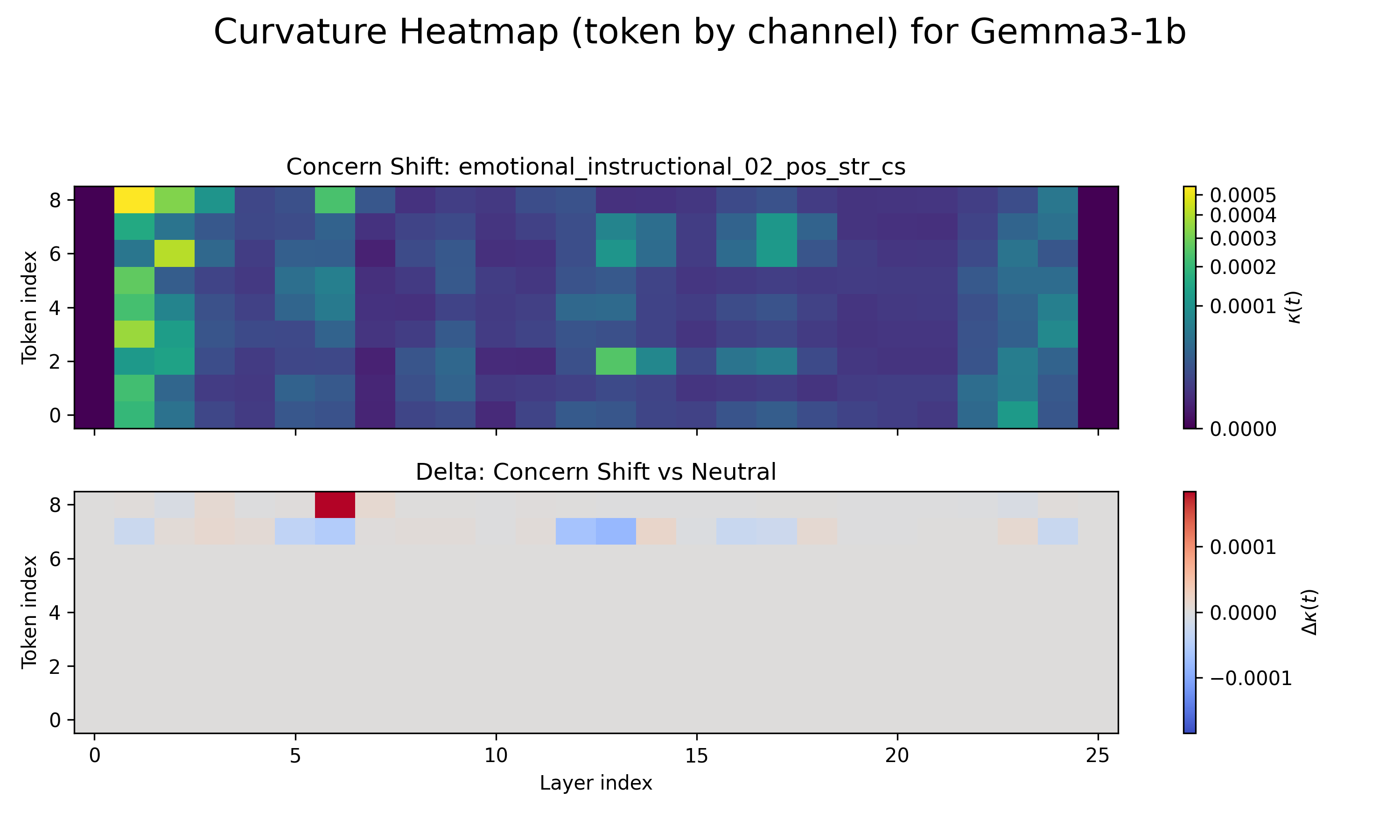}

\begin{figure}[!h]
\caption{Neutral and Delta Curvature heatmaps for each variant of a single prompt using Gemma3-1b}
\end{figure}

\includegraphics[width=0.48\textwidth,height=\textheight]{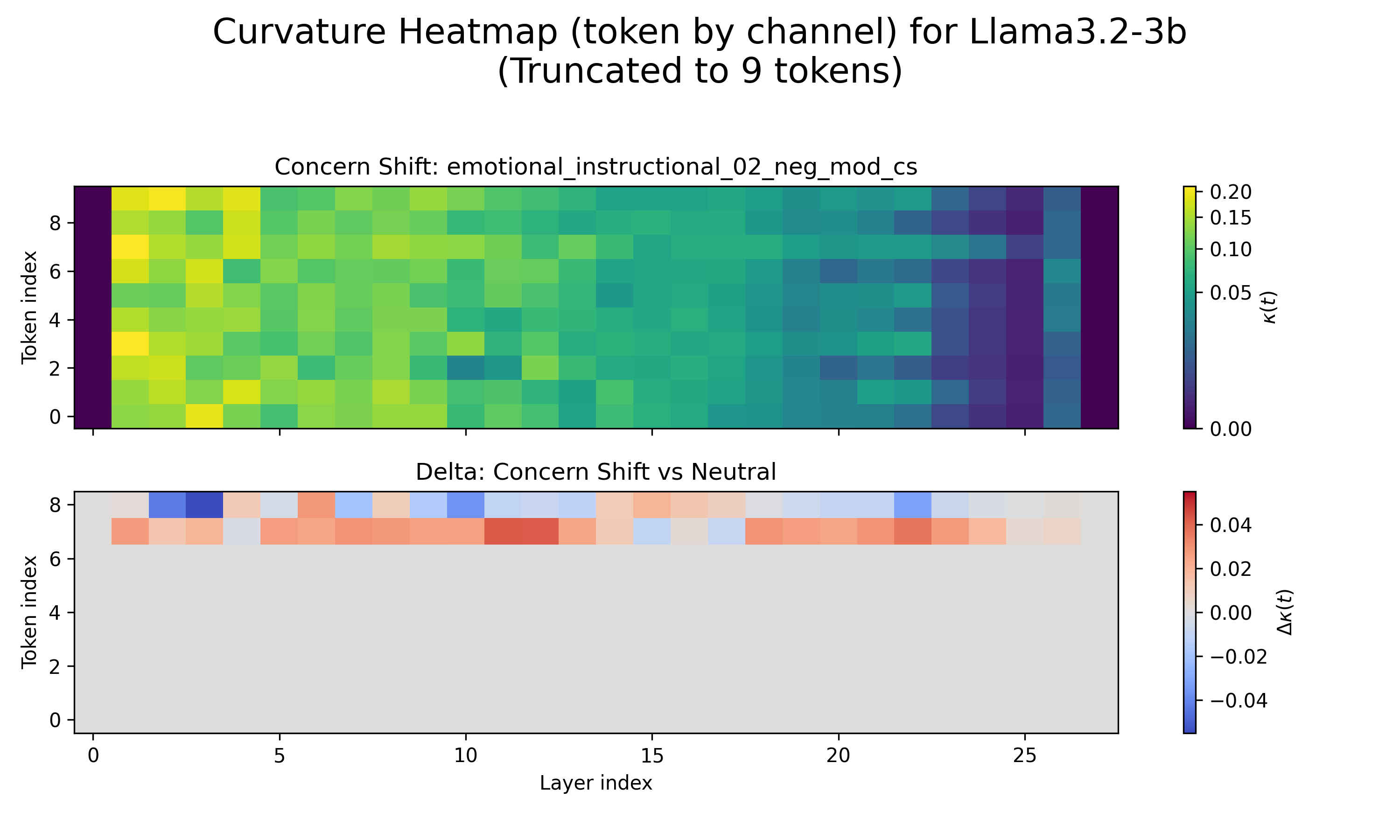}
\includegraphics[width=0.48\textwidth,height=\textheight]{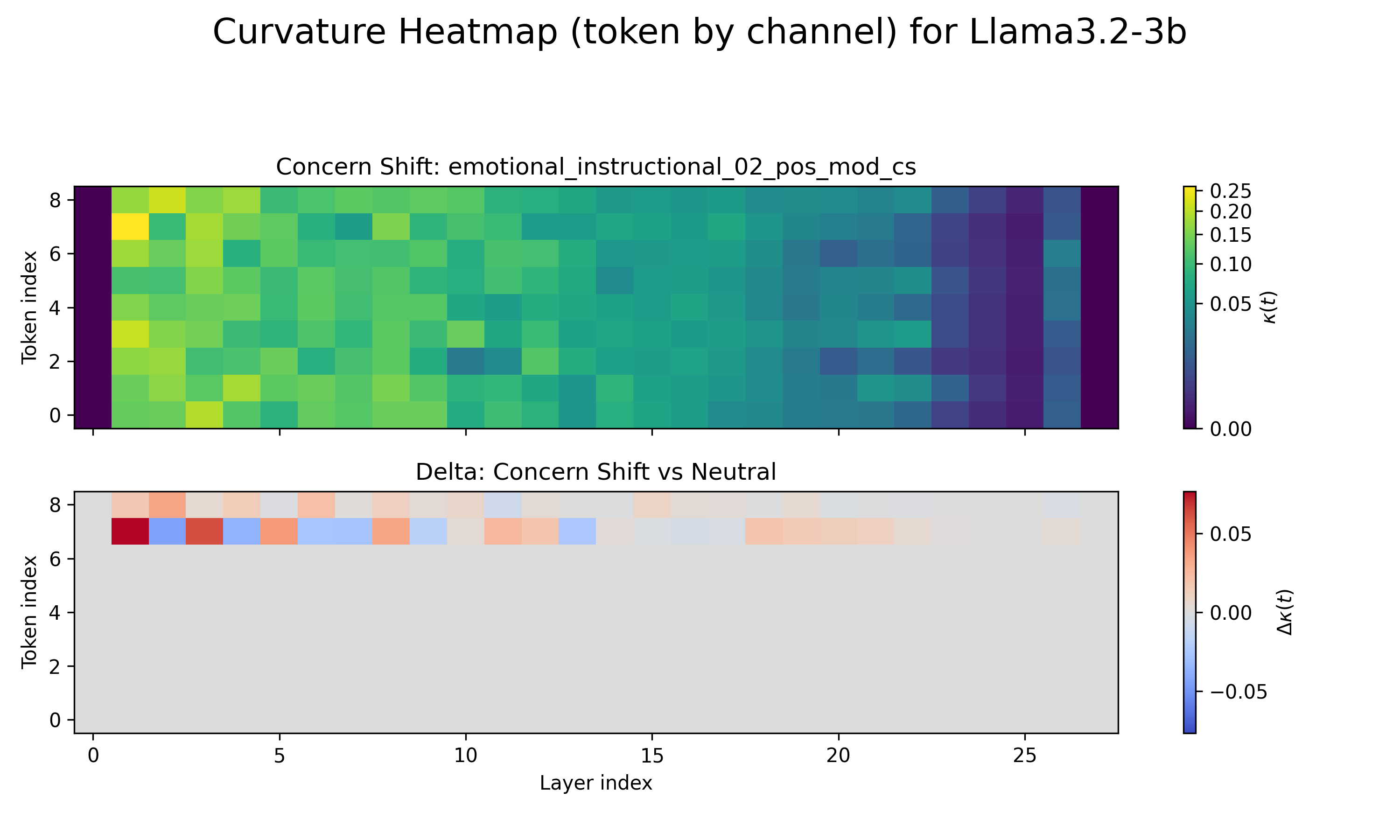}

\includegraphics[width=0.48\textwidth,height=\textheight]{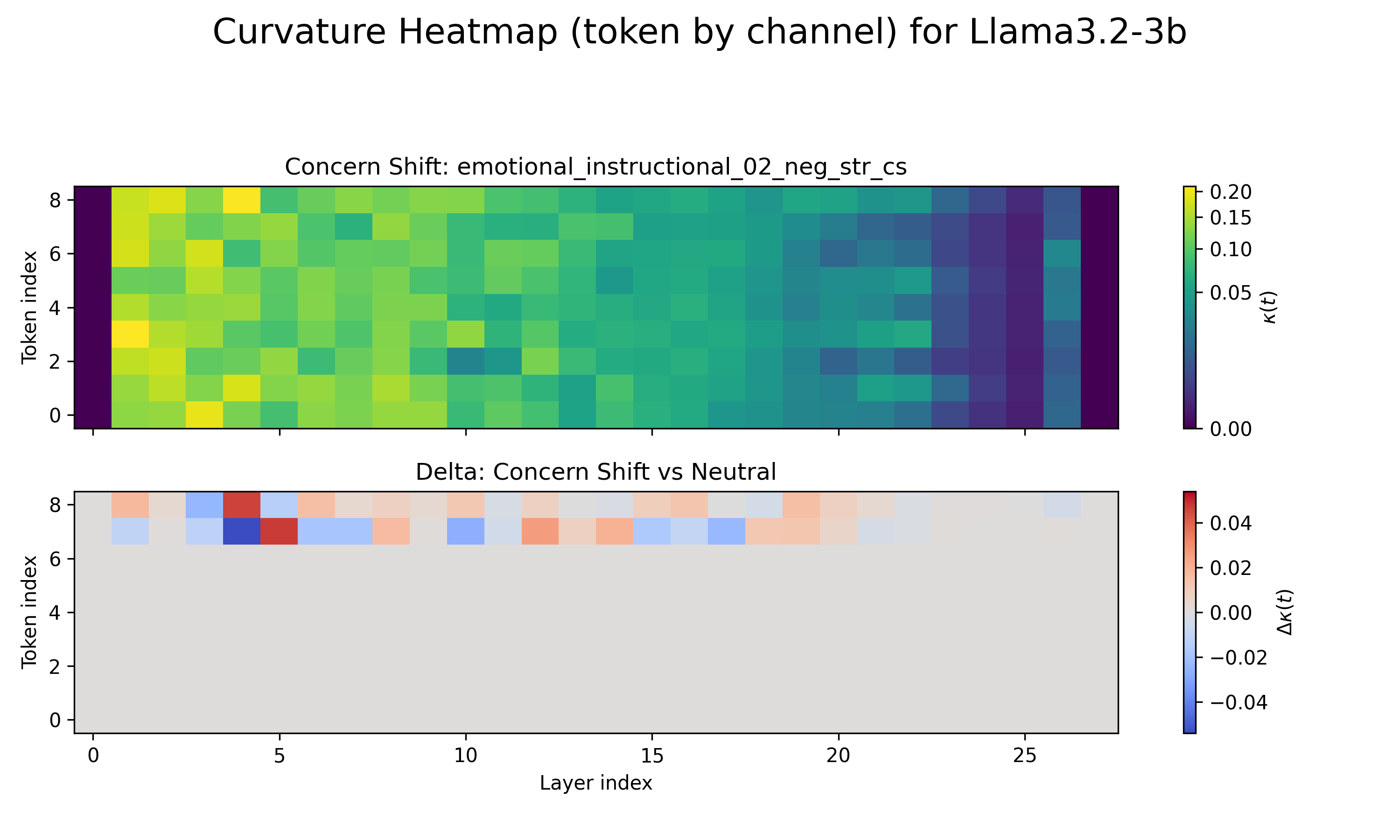}
\includegraphics[width=0.48\textwidth,height=\textheight]{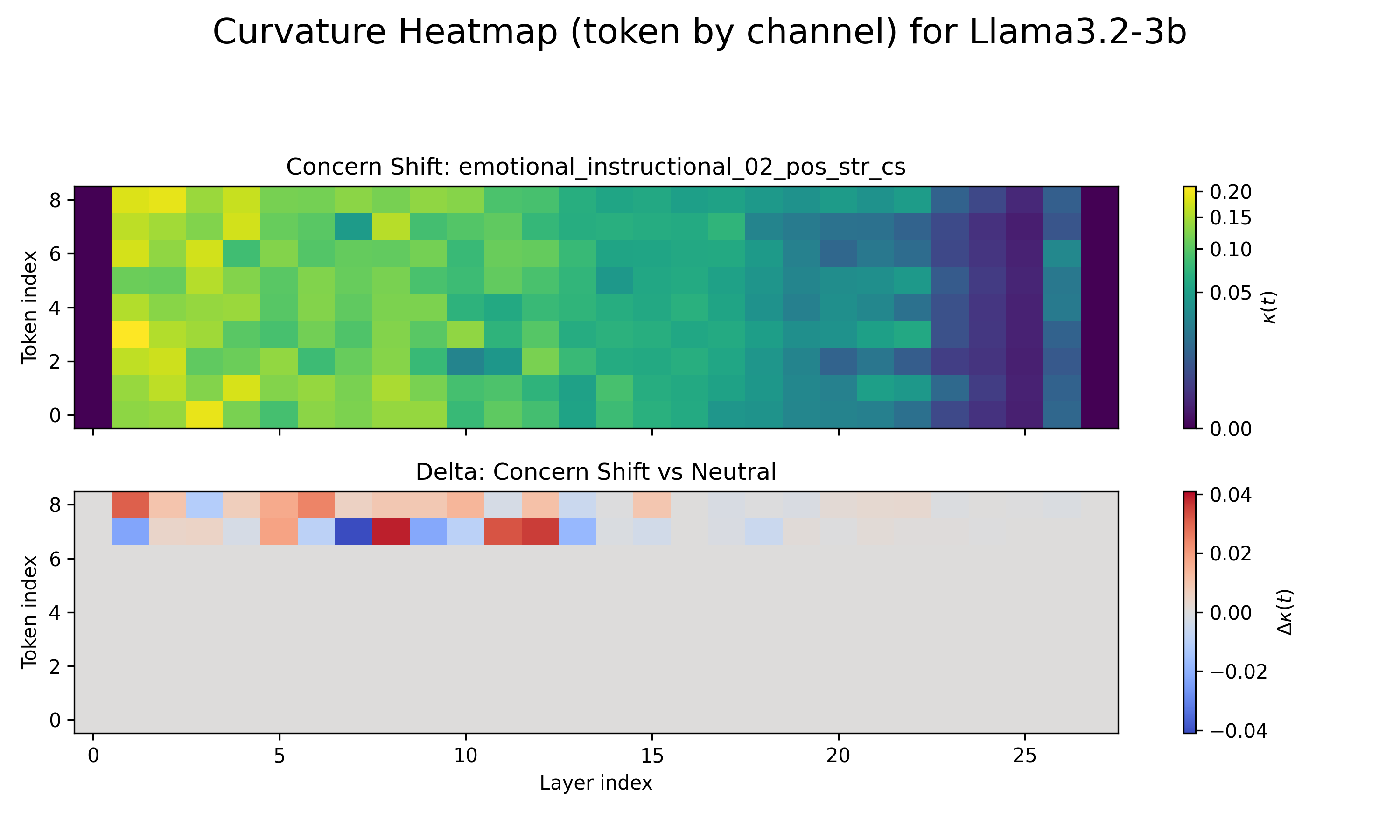}

\begin{figure}[!h]
\caption{Neutral and Delta Curvature heatmaps for each variant of a single prompt using LLaMA3.2-3b}
\end{figure}

\includegraphics[width=0.48\textwidth,height=\textheight]{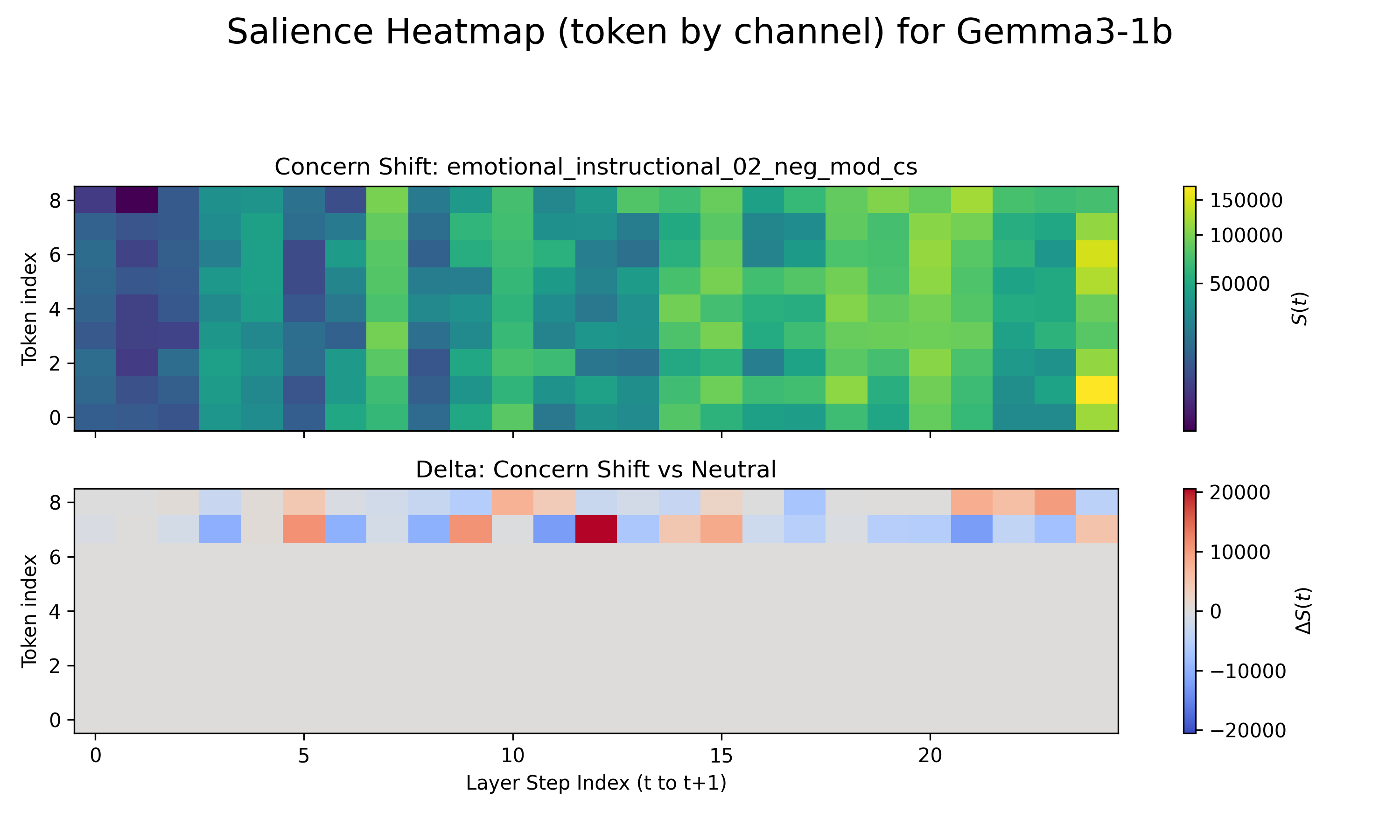}
\includegraphics[width=0.48\textwidth,height=\textheight]{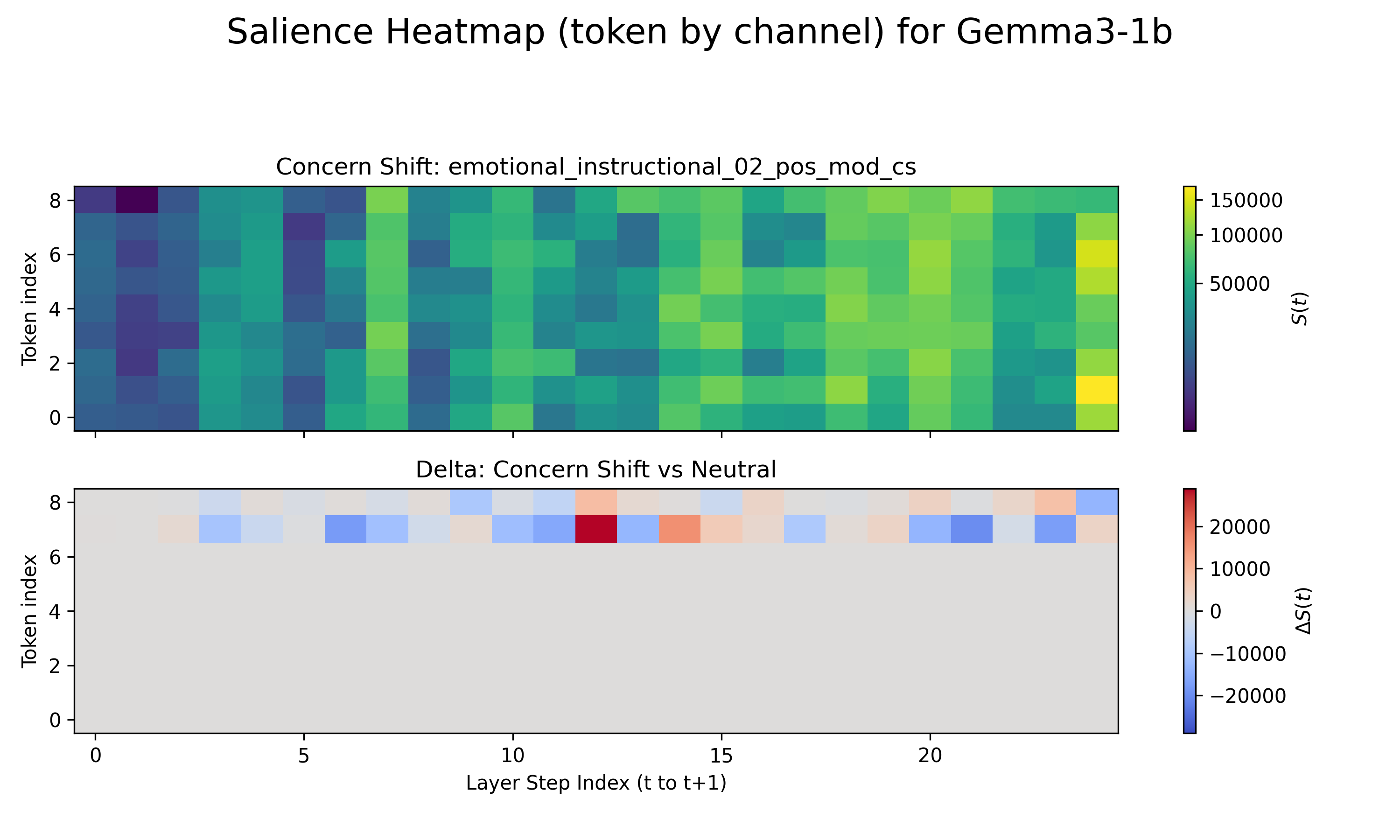}

\includegraphics[width=0.48\textwidth,height=\textheight]{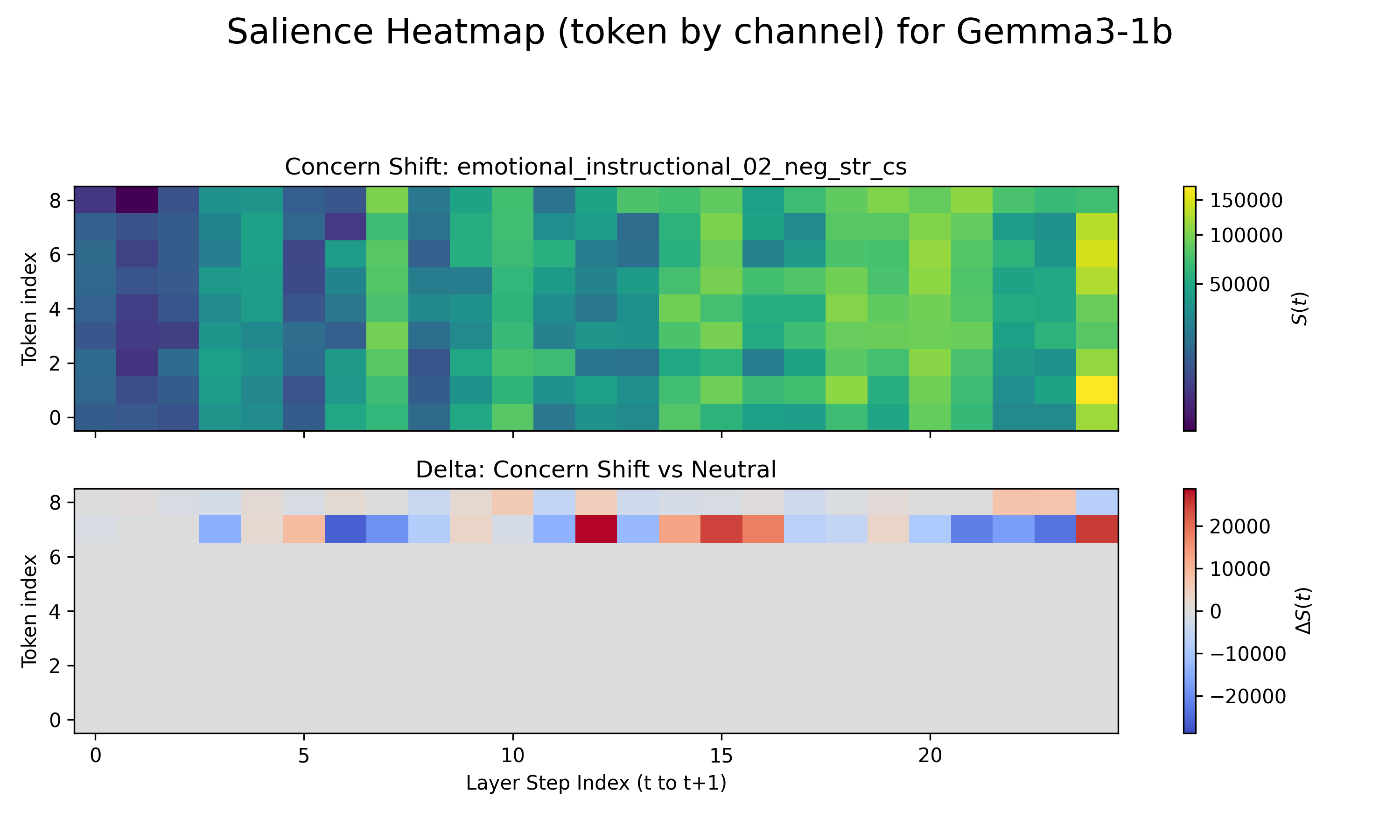}
\includegraphics[width=0.48\textwidth,height=\textheight]{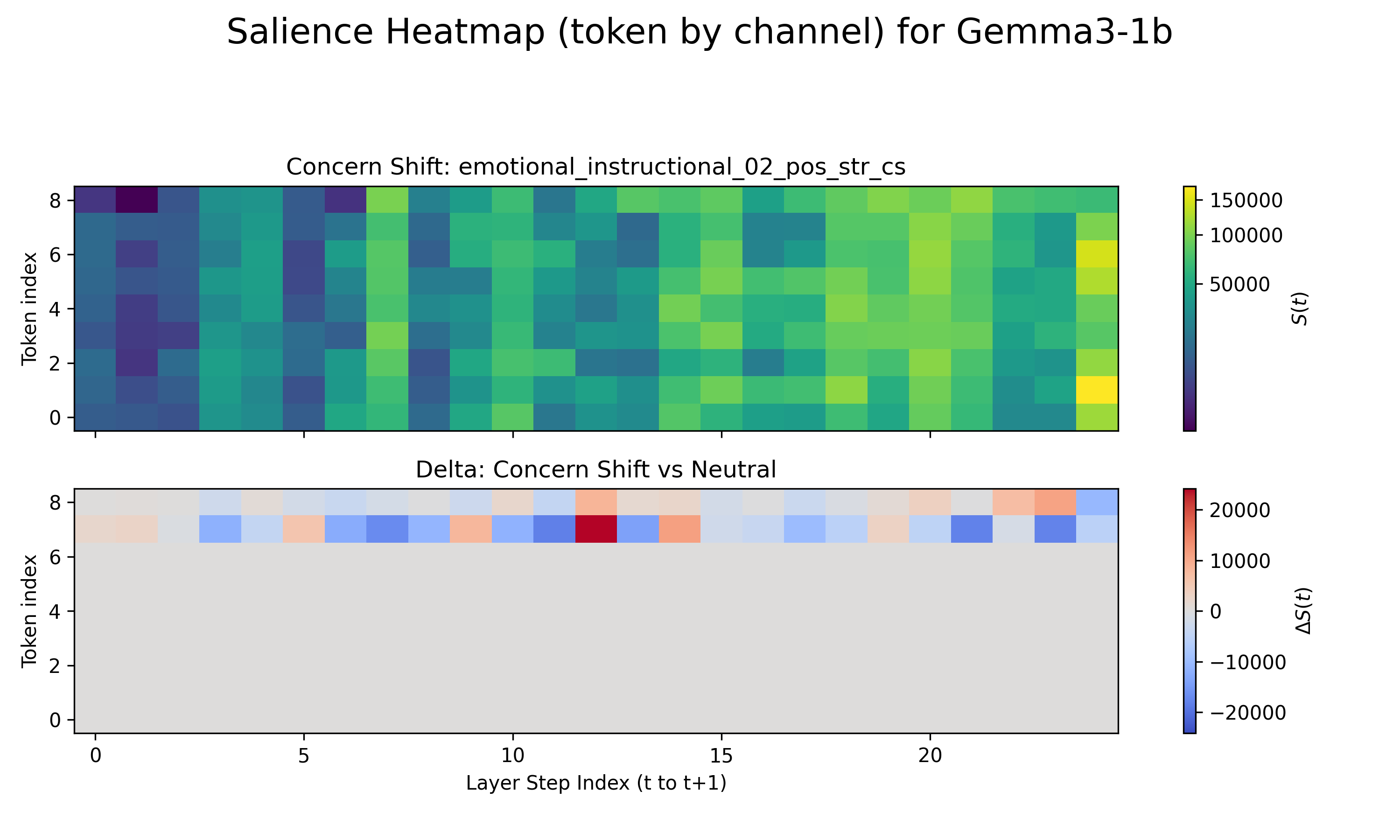}

\begin{figure}[!h]
\caption{Neutral and Delta Salience heatmaps for each variant of a single prompt using Gemma3-1b}
\end{figure}

\includegraphics[width=0.48\textwidth,height=\textheight]{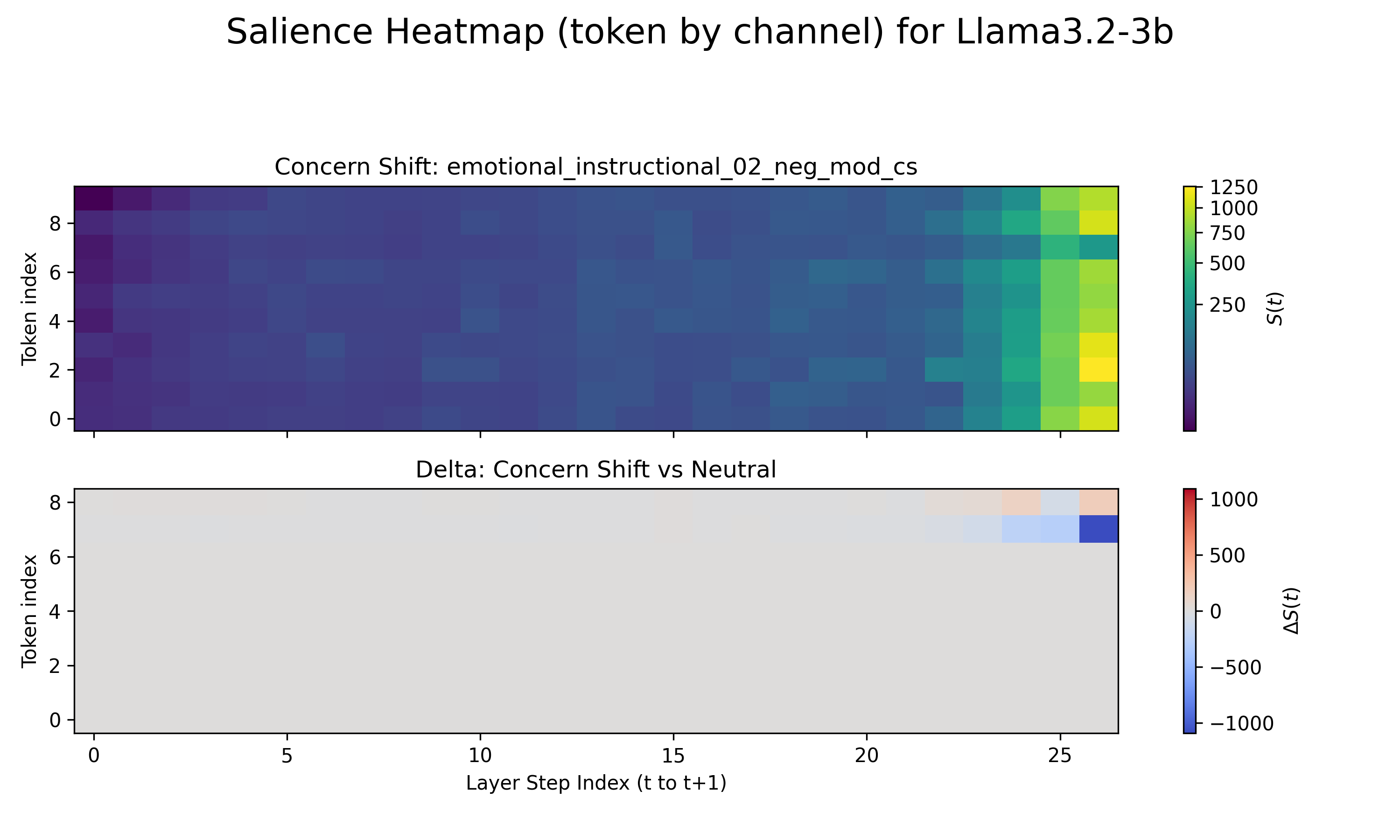}
\includegraphics[width=0.48\textwidth,height=\textheight]{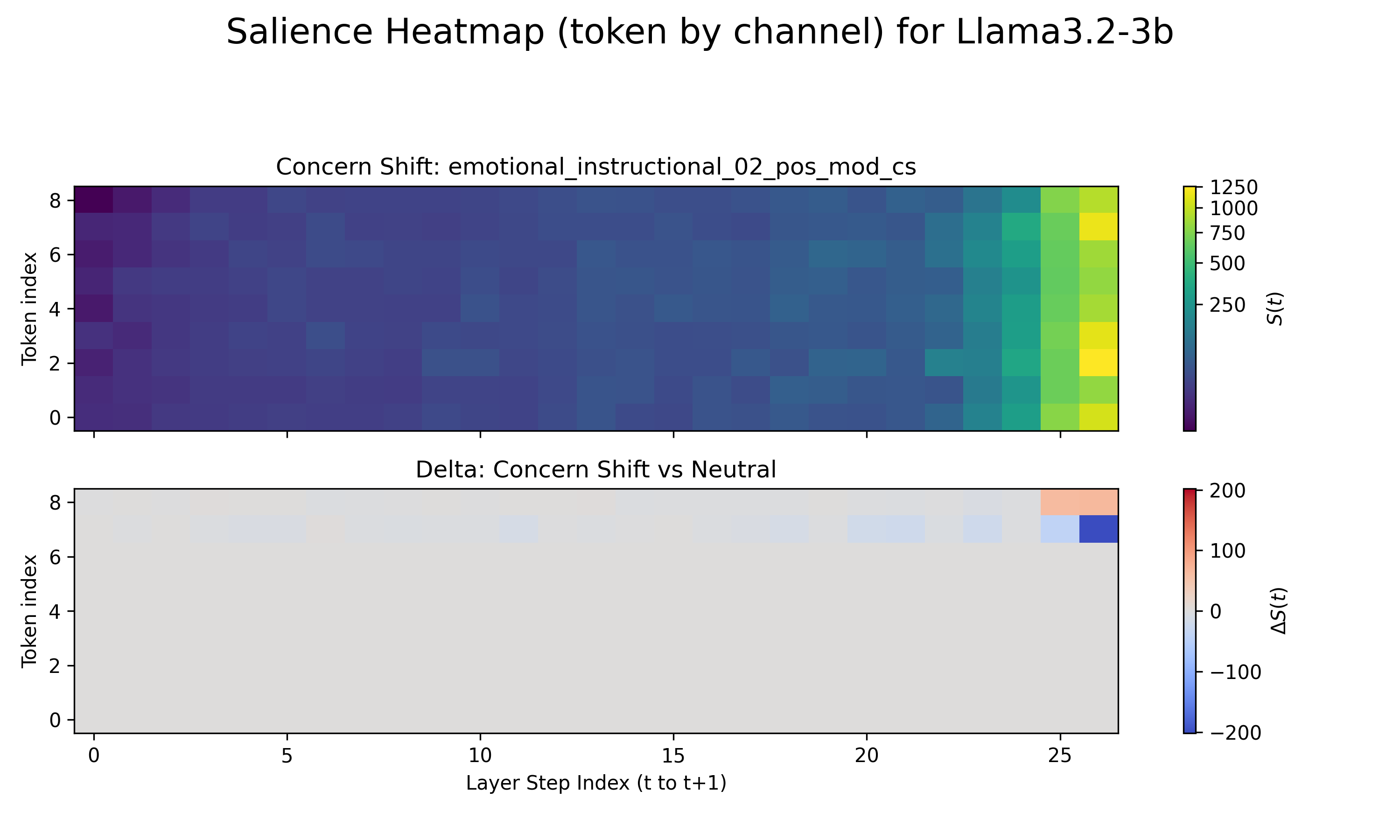}

\includegraphics[width=0.48\textwidth,height=\textheight]{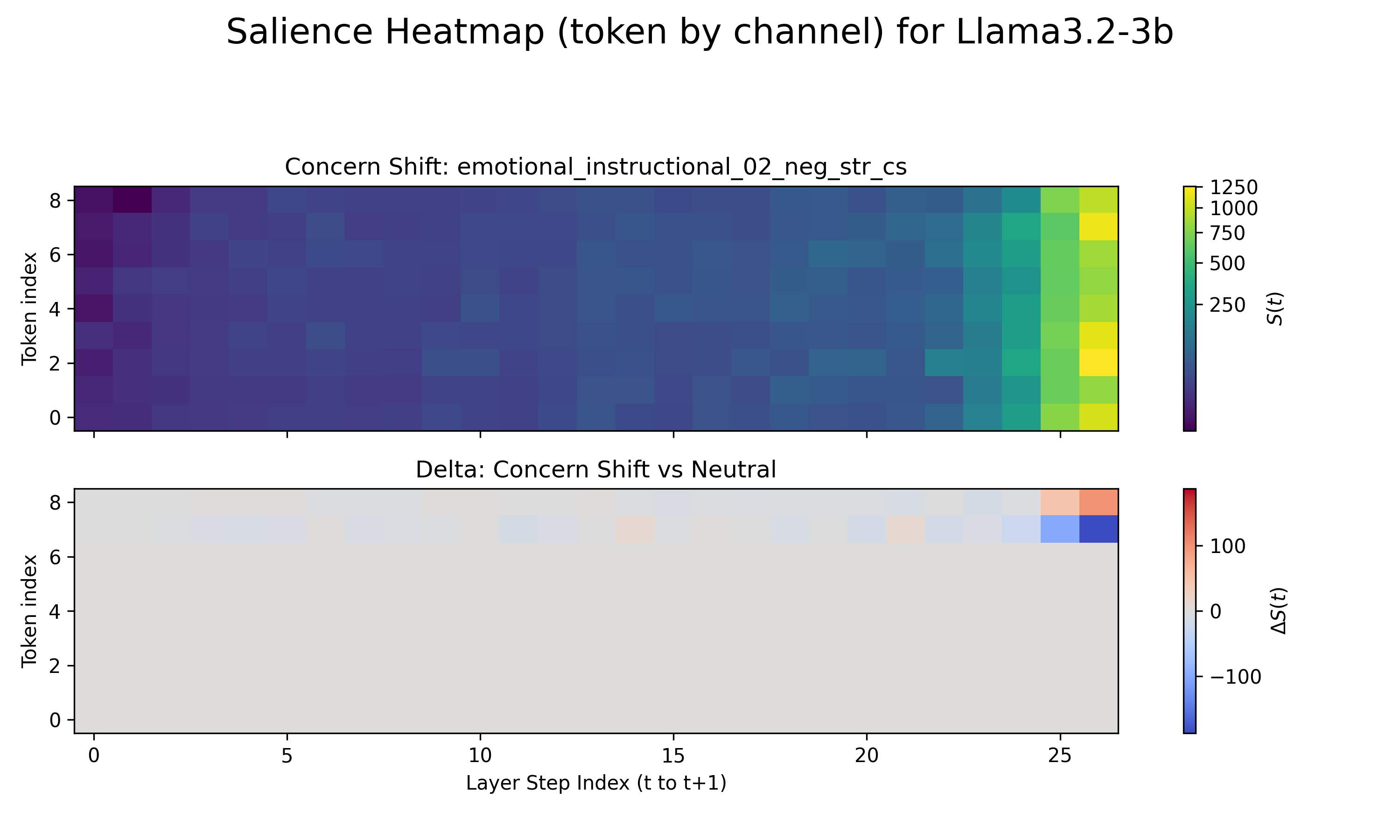}
\includegraphics[width=0.48\textwidth,height=\textheight]{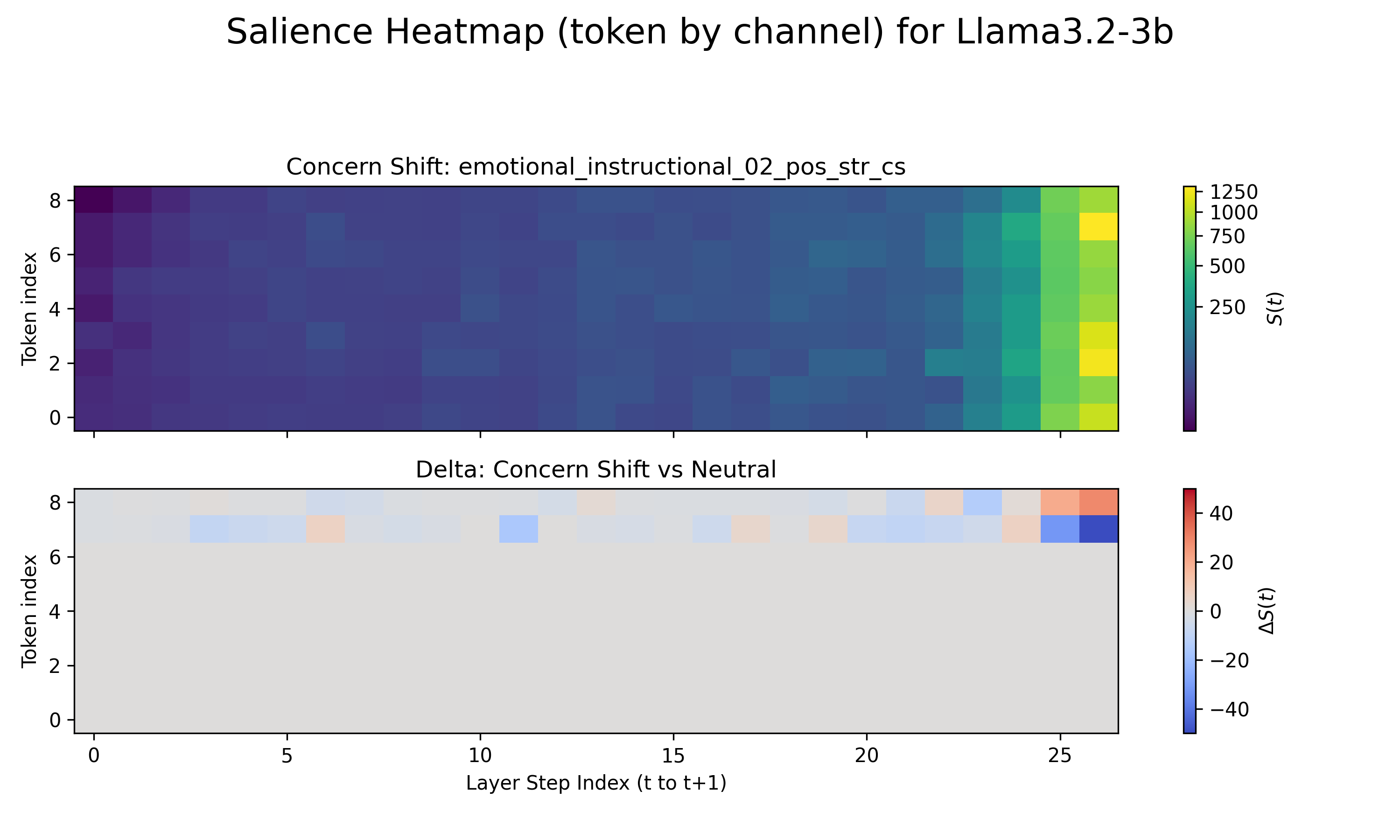}

\begin{figure}[!h]
\caption{Neutral and Delta Salience heatmaps for each variant of a single prompt using LLaMA3.2-3b}
\end{figure}

\textbf{Table 3 Summary of Path-Curvature Metrics}

\begin{longtable}[]{@{}
  >{\raggedright\arraybackslash}p{(\columnwidth - 8\tabcolsep) * \real{0.1129}}
  >{\raggedright\arraybackslash}p{(\columnwidth - 8\tabcolsep) * \real{0.1613}}
  >{\raggedright\arraybackslash}p{(\columnwidth - 8\tabcolsep) * \real{0.2742}}
  >{\raggedright\arraybackslash}p{(\columnwidth - 8\tabcolsep) * \real{0.2581}}
  >{\raggedright\arraybackslash}p{(\columnwidth - 8\tabcolsep) * \real{0.1935}}@{}}
\toprule\noalign{}
\begin{minipage}[b]{\linewidth}\raggedright
Model
\end{minipage} & \begin{minipage}[b]{\linewidth}\raggedright
Variant
\end{minipage} & \begin{minipage}[b]{\linewidth}\raggedright
Mean 3-point \(\kappa\)
\end{minipage} & \begin{minipage}[b]{\linewidth}\raggedright
Max 3-point \(\kappa\)
\end{minipage} & \begin{minipage}[b]{\linewidth}\raggedright
Layer\(_{\text{max}}\)
\end{minipage} \\
\midrule\noalign{}
\endhead
\bottomrule\noalign{}
\endlastfoot
Gemma & neutral & 0.0000372 & 0.0003122 & 5.4 \\
Gemma & neg\_mod & 0.0000366 & 0.0003037 & 5.6 \\
Gemma & pos\_mod & 0.0000372 & 0.0003168 & 5.3 \\
Gemma & neg\_str & 0.0000377 & 0.0003199 & 5.6 \\
Gemma & pos\_str & 0.0000370 & 0.0003113 & 5.2 \\
LLaMA & neutral & 0.0743119 & 0.1760759 & 3.1 \\
LLaMA & neg\_mod & 0.0744599 & 0.1770283 & 2.9 \\
LLaMA & pos\_mod & 0.0745456 & 0.1785080 & 3.0 \\
LLaMA & neg\_str & 0.0749832 & 0.1785855 & 3.0 \\
LLaMA & pos\_str & 0.0747999 & 0.1768842 & 2.9 \\
\end{longtable}

\textbf{Key observations:}

\begin{itemize}
\tightlist
\item
  \textbf{LLaMA exhibits early, high-magnitude semantic curvature}, with
  peak values concentrated around layer 3, and broad curvature sustained
  through mid-depth layers. This reflects fast, distributed integration
  of concern information.
\item
  \textbf{Gemma shows weak, shallow curvature} concentrated around
  layers 5-6, indicating limited semantic bending and fast flattening of
  signal.
\item
  \textbf{Concern-shifted prompts} (especially strong polarity) elevate
  curvature in both models, but \textbf{neutral prompts} also produce
  high \(\kappa\) in LLaMA, suggesting consistent trajectory shaping
  even in low-salience conditions.
\end{itemize}

In many cases, curvature peaks occur shortly after concern tokens and
are consistent with the discrete curvature criterion introduced in
Appendix C.

We call a layer `early' if it lies in the first quartile of the network
depth (layers 1-7 in our 28-layer LLaMA, layers 1-6 in 26-layer Gemma).
Curvature onsets in this early band for both models, but only LLaMA
sustains high curvature through mid-depth (layers 8-14), so its
layer-of-max statistic lands at around 3 even though the heatmap shows a
broad high-\(\kappa\) region.

\textbf{Note:} \emph{The layer-of-max metrics in Table 3 reflect average
scalar curvature across tokens, whereas the heatmaps depict full
token-layer distributions. In LLaMA, curvature often begins early but
remains visually sustained across mid-depth layers - revealing a broader
narrative arc than a single summary statistic can capture.}

\hypertarget{inter-metric-relationships}{%
\subsubsection{4.2 Inter-metric
Relationships}\label{inter-metric-relationships}}

Across all 100 prompt-variant pairs, we observe a strong anticorrelation
between \textbf{cosine similarity} and \textbf{layer-wise Euclidean
deviation}, with Pearson correlation coefficients of \(r = -0.95\)
(Gemma) and \(r = -0.98\) (LLaMA). This supports the interpretation that
concern-shifted prompts tend to \textbf{reorient} the model's internal
representation more than they \textbf{displace} it (a \emph{pivot rather
than a drift}), indicating semantic reconfiguration without runaway norm
inflation.

In contrast, we find only a weak or negligible correlation between
\textbf{path curvature} \(\kappa_i\) and \textbf{layer-\(\Delta\)} (the
average angular deviation between matched directional steps), with
Pearson \(r = -0.28\) (Gemma) and \(r = -0.01\) (LLaMA). This suggests
that while both metrics index semantic reorientation, they may capture
\textbf{distinct geometric behaviours}:

\begin{quote}
Curvature reflects \textbf{local inflection} within token trajectories,
while direction deviation aggregates \textbf{global path shape} across
the full residual arc.
\end{quote}

\textbf{Table 4 Summary of Metric Correlation}

\begin{longtable}[]{@{}
  >{\raggedright\arraybackslash}p{(\columnwidth - 8\tabcolsep) * \real{0.1375}}
  >{\raggedright\arraybackslash}p{(\columnwidth - 8\tabcolsep) * \real{0.2375}}
  >{\raggedright\arraybackslash}p{(\columnwidth - 8\tabcolsep) * \real{0.2500}}
  >{\raggedright\arraybackslash}p{(\columnwidth - 8\tabcolsep) * \real{0.2375}}
  >{\raggedright\arraybackslash}p{(\columnwidth - 8\tabcolsep) * \real{0.1375}}@{}}
\toprule\noalign{}
\begin{minipage}[b]{\linewidth}\raggedright
Model
\end{minipage} & \begin{minipage}[b]{\linewidth}\raggedright
Cosine vs Euclidean (r)
\end{minipage} & \begin{minipage}[b]{\linewidth}\raggedright
Curvature vs Direction (r)
\end{minipage} & \begin{minipage}[b]{\linewidth}\raggedright
Salience vs Curvature (r)
\end{minipage} & \begin{minipage}[b]{\linewidth}\raggedright
Prompt Count
\end{minipage} \\
\midrule\noalign{}
\endhead
\bottomrule\noalign{}
\endlastfoot
Gemma3-1b & -0.9524 & -0.2779 & -0.5562 & 100 \\
LLaMA3.2-3b & -0.9784 & -0.0061 & -0.8931 & 100 \\
\end{longtable}

As shown in Table 4, we also observe a strong \textbf{anticorrelation
between salience and curvature}, with \(r = -0.56\) (Gemma) and
\(r = -0.89\) (LLaMA). Conceptually, \textbf{salience} (total
representational movement) and \textbf{curvature} (degree of directional
reorientation) are distinct geometric properties:

\begin{quote}
A model can move far without turning sharply, or turn sharply without
covering much distance.
\end{quote}

However, the strong negative correlation observed for LLaMA indicates a
\textbf{systematic behavioural tendency}:

\begin{quote}
When token trajectories are highly curved, they tend to be shorter in
total length; and when total movement is high, the trajectory is
typically straighter.
\end{quote}

This inverse relationship suggests a kind of \textbf{representational
trade-off}: LLaMA often seems to prioritise either \textbf{distance} or
\textbf{reorientation}, but not both to an extreme degree within the
same prompt group. This may reflect an internal efficiency mechanism - a
kind of ``representational budget'' that balances semantic effort with
semantic precision.

In Gemma, this trade-off is present but less pronounced, consistent with
its generally shallower curvature and lower overall salience. The weaker
correlation suggests a more variable relationship between reorientation
and effort, or simply less pronounced geometric specialisation.

These geometric patterns emerge only in the residual stream, the
\textbf{unique cumulative path} of semantic inference inside transformer
models. As detailed in Appendix A (section A.6), attention and MLP
layers act as \textbf{semantic lenses} - bending and redirecting token
trajectories based on relational and nonlinear context. Their outputs
are added to the residual stream, shaping its curvature.

\begin{quote}
The model may bend its internal trajectory without making large semantic
leaps.
\end{quote}

Since attention and MLP outputs are the primary forces applied to the
residual stream, these curvature and salience patterns must originate in
their lensing effects.

\hfill\break

Thus, these heatmaps offer a \textbf{window into the emergent geometry
of meaning}:

\begin{quote}
The attention and MLP forces remain invisible, but their
\emph{integrated effect} (curvature and salience), can be clearly seen.
\end{quote}

Once seen, these patterns become difficult to unsee.

Rather than treating curvature and directional deviation as proxies for
the same phenomenon, we interpret them as \textbf{complementary
signals}. \textbf{Curvature} captures \emph{where and how sharply} the
model bends its internal trajectory - often in response to localised
semantic inflections induced by concern-shifted tokens. In contrast,
\textbf{directional deviation (layer-\(\Delta\))} reflects \emph{how far
the representation pivots overall}, accumulating changes across layers.

\begin{quote}
\begin{quote}
\emph{Curvature captures where and how sharply the model bends its
internal trajectory}
\end{quote}
\end{quote}

This distinction aligns with the \textbf{Semantic Lens} view introduced
in Appendix A:

\begin{quote}
Attention and MLP block layers apply localised semantic forces at each
layer, producing curvature in the residual stream.
\end{quote}

The token-by-layer heatmaps make this clear - sharp bends are often
concentrated around specific layers and token positions. These local
twists may not significantly change the overall trajectory arc but still
encode important semantic reorientations.

\begin{quote}
\textbf{Interpretive claim:}

\begin{quote}
If curvature \(\kappa_i\) encodes \textbf{semantic reorientation}, and
salience \(S(t)\) encodes \textbf{semantic effort}, then their joint
distribution across tokens and layers reveals \textbf{how the model
prioritises and navigates internal meaning} under pressure from latent
concern.
\end{quote}
\end{quote}

\hypertarget{quantifying-concern-shift-effects-across-prompt-variants}{%
\subsubsection{4.3 Quantifying Concern-Shift Effects Across Prompt
Variants}\label{quantifying-concern-shift-effects-across-prompt-variants}}

To complement the geometric heatmaps in Section 4.1, we now present a
summary of concern-shift (CS) effects across all 20 prompt sets,
comparing ``moderate'' and ``strong'' variants in both positive and
negative polarities. While earlier sections focused on individual
visualisations, this section quantifies how reliably these concern
manipulations induce geometric changes in the residual stream, using
both curvature and salience metrics.

We report mean absolute delta values for each prompt variant, computed
by subtracting the neutral control metric and measuring the layer-wise
magnitude of change across all tokens. For each model, we test whether
these deltas increase with concern strength, and whether the observed
effects are statistically significant across prompts.

\hypertarget{summary-statistics}{%
\paragraph{\texorpdfstring{Summary Statistics:\\
}{Summary Statistics: }}\label{summary-statistics}}

\hfill\break
Table 5 presents delta analysis results comparing Moderate vs Strong
concern-shifts.

This data shows that:

\begin{itemize}
\tightlist
\item
  \textbf{LLaMA3.2-3b} exhibits consistent and statistically significant
  increases in both curvature and salience from moderate to strong
  concern in positive prompts.
\item
  \textbf{Gemma3-1b}, while showing non-zero deltas, does not exhibit
  significant scaling with concern strength.
\item
  Negative polarity prompts in both models show more variability and
  less consistent statistical separation.
\end{itemize}

\textbf{Table 5 Summary of Delta Magnitudes and Significance}

\begin{longtable}[]{@{}
  >{\raggedright\arraybackslash}p{(\columnwidth - 12\tabcolsep) * \real{0.1899}}
  >{\raggedright\arraybackslash}p{(\columnwidth - 12\tabcolsep) * \real{0.1266}}
  >{\raggedright\arraybackslash}p{(\columnwidth - 12\tabcolsep) * \real{0.1013}}
  >{\raggedright\arraybackslash}p{(\columnwidth - 12\tabcolsep) * \real{0.1519}}
  >{\raggedright\arraybackslash}p{(\columnwidth - 12\tabcolsep) * \real{0.1519}}
  >{\raggedright\arraybackslash}p{(\columnwidth - 12\tabcolsep) * \real{0.1139}}
  >{\raggedright\arraybackslash}p{(\columnwidth - 12\tabcolsep) * \real{0.1646}}@{}}
\toprule\noalign{}
\begin{minipage}[b]{\linewidth}\raggedright
Model
\end{minipage} & \begin{minipage}[b]{\linewidth}\raggedright
Metric
\end{minipage} & \begin{minipage}[b]{\linewidth}\raggedright
Polarity
\end{minipage} & \begin{minipage}[b]{\linewidth}\raggedright
Mean \(\Delta\) (Mod)
\end{minipage} & \begin{minipage}[b]{\linewidth}\raggedright
Mean \(\Delta\) (Str)
\end{minipage} & \begin{minipage}[b]{\linewidth}\raggedright
Ratio S/M
\end{minipage} & \begin{minipage}[b]{\linewidth}\raggedright
p(Str \textgreater{} Mod)\(_t\)
\end{minipage} \\
\midrule\noalign{}
\endhead
\bottomrule\noalign{}
\endlastfoot
Gemma3-1b & Curvature & Positive & 0.0000 & 0.0000 & 1.59 & 0.127 \\
Gemma3-1b & Curvature & Negative & 0.0000 & 0.0000 & 1.49 & 0.366 \\
Gemma3-1b & Salience & Positive & 3680.0000 & 4520.0000 & 1.51 &
0.152 \\
Gemma3-1b & Salience & Negative & 4760.0000 & 4990.0000 & 1.48 &
0.389 \\
LLaMA3.2-3b & Curvature & Positive & 0.0063 & 0.0085 & 1.38 &
\textbf{0.006} \\
LLaMA3.2-3b & Curvature & Negative & 0.0067 & 0.0072 & 1.24 & 0.265 \\
LLaMA3.2-3b & Salience & Positive & 7.3800 & 11.1000 & 1.80 &
\textbf{0.016} \\
LLaMA3.2-3b & Salience & Negative & 8.0300 & 8.7300 & 1.59 & 0.342 \\
\end{longtable}

\hypertarget{visual-summary}{%
\paragraph{\texorpdfstring{Visual Summary:\\
}{Visual Summary: }}\label{visual-summary}}

\hfill\break
Figures 7 and 8 provide a per-prompt visualisation of these deltas. Each
subplot shows the mean absolute delta in curvature or salience for both
``moderate'' (blue circle) and ``strong'' (red cross) CS variants.
Prompts are sorted by total delta magnitude to emphasise those with the
most pronounced geometric effects.

These plots reveal:

\begin{itemize}
\tightlist
\item
  \textbf{In LLaMA}, strong variants often exceed moderate ones for both
  metrics, particularly in prompts such as ``Perspective Advice 08'' and
  ``Logical If Then 11''.
\item
  \textbf{In Gemma}, the pattern is noisier. Some prompts show minimal
  difference or even larger deltas for moderate concern.
\end{itemize}

Overall, this analysis supports the interpretation that concern-shifted
prompts do not merely alter model output, but reshape internal
representation in a graded, model-sensitive manner. Stronger concern
often correlates with larger semantic trajectory shifts in LLaMA, and to
a lesser extent in Gemma. These shifts are evident both in raw metric
magnitudes and in structured per-prompt trends.

\begin{quote}
In all cases, concern-shifted prompts create curvature and salience
deviations from the control prompts.
\end{quote}

\includegraphics[width=0.95\textwidth,height=\textheight]{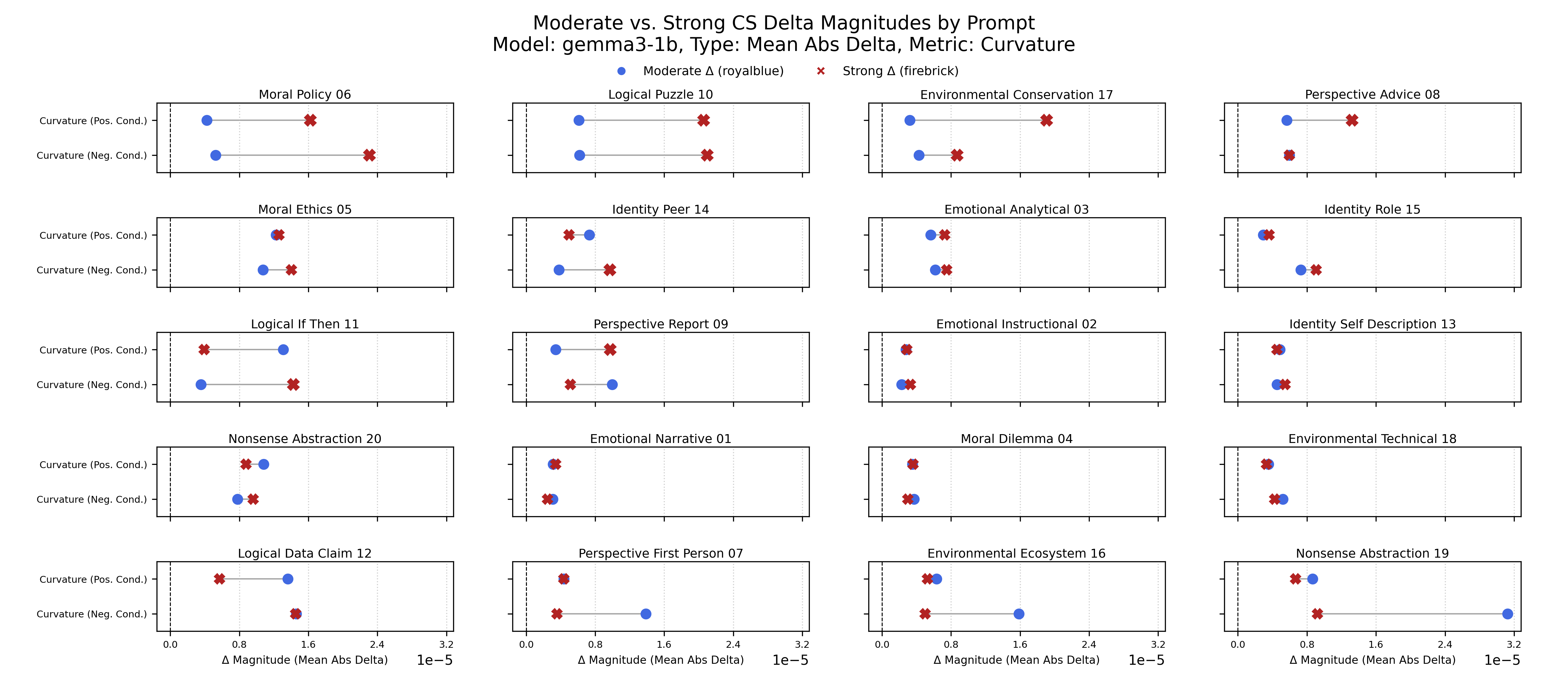}
\includegraphics[width=0.95\textwidth,height=\textheight]{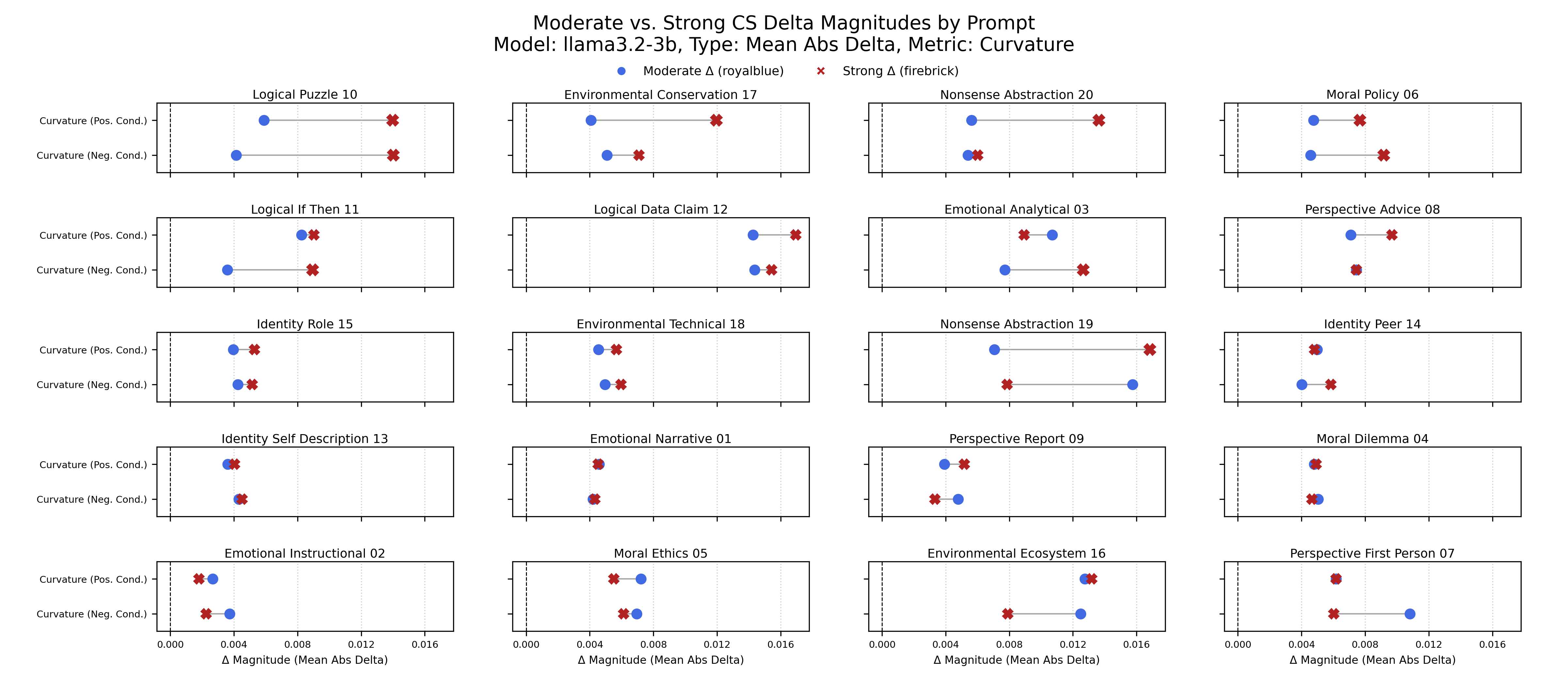}

\begin{figure}[!h]
\caption{Mean Curvature Delta plots by prompt using Gemma3-1b}
\end{figure}

\includegraphics[width=0.95\textwidth,height=\textheight]{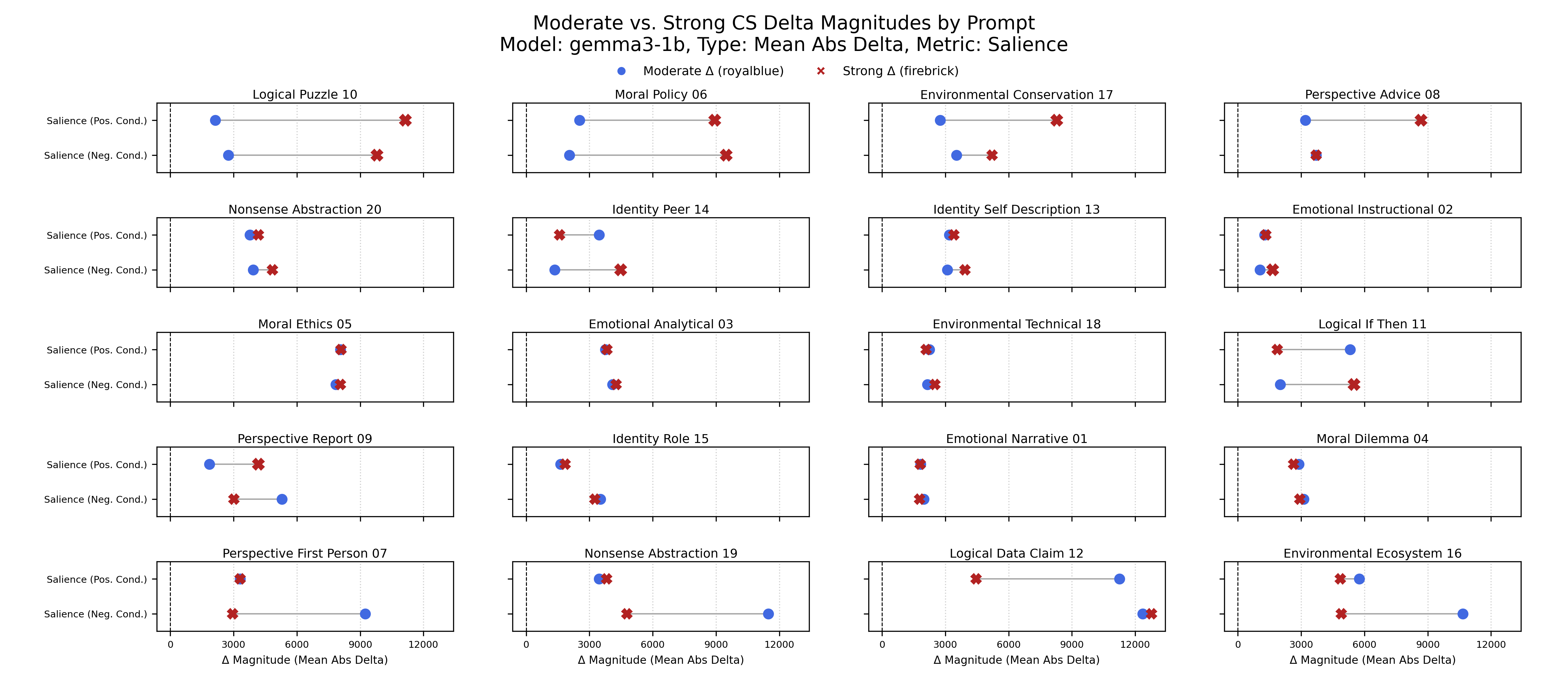}
\includegraphics[width=0.95\textwidth,height=\textheight]{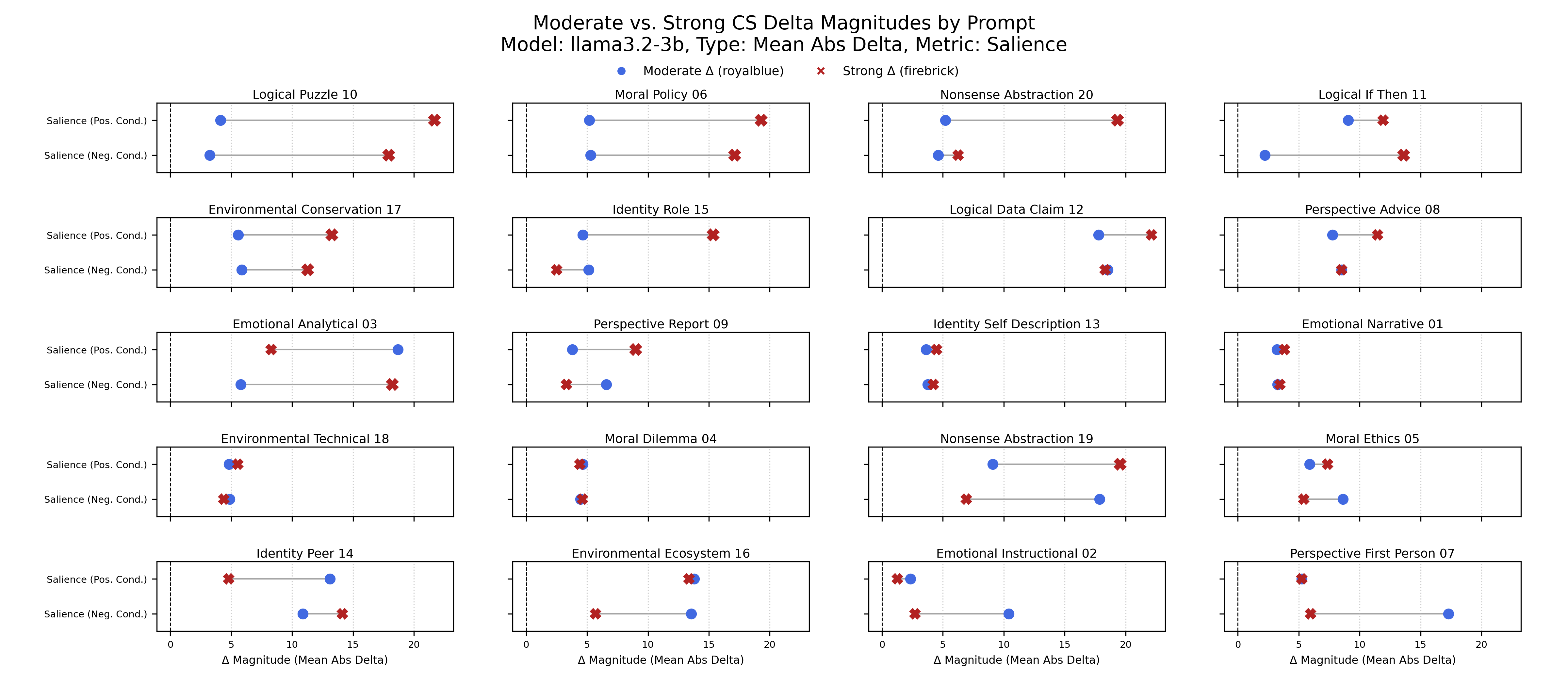}

\begin{figure}[!h]
\caption{Mean Salience Delta plots by prompt using LLaMA3.2-3b}
\end{figure}

\hypertarget{discussion}{%
\subsection{5 Discussion}\label{discussion}}

The results of this study highlight the value of \textbf{residual stream
curvature} as an emergent geometric signal of concern-sensitive
processing in transformer models. Unlike surface-level metrics or
attribution-based scores, curvature and salience offer a
\textbf{dynamic, trajectory-level view} of how models integrate,
differentiate, and abstract contextual meaning in response to semantic
concern.

This discussion reaffirms that \textbf{curvature is a real,
geometry-grounded signal} - whereas low-dimensional projection artefacts
are not. The shape of the residual stream offers a fertile new
diagnostic for LLM Interpretability.

\hypertarget{key-takeaways}{%
\subsubsection{5.1 Key Takeaways}\label{key-takeaways}}

Our analysis yields four principal findings:

\begin{enumerate}
\def\labelenumi{\arabic{enumi}.}
\item
  \textbf{Low-dimensional projections distort curvature geometry.} An
  earlier audit of our initial results that utilised UMAP to project the
  data into a lower-dimensional space showed median angle correlations
  of 0.47 (Gemma) and -0.25 (LLaMA), far below the fidelity threshold
  (\(r \ge 0.80\)), rendering UMAP-based curvature measurements
  unreliable. This reinforces our conclusion that \textbf{native-space
  analysis is essential}.
\item
  \textbf{Path curvature \(\kappa_i\) unifies and sharpens geometric
  insights.} The discrete 3-point curvature metric \(\kappa_i\) captures
  all qualitative effects hinted at by initial projection visuals -
  while remaining mathematically grounded, semantically aligned, and
  independent of dimensionality reduction.
\item
  \textbf{Curvature tracks salience - but its expression depends on
  architecture.} Gemma3-1b exhibits an early, shallow, localised bend,
  while LLaMA3.2-3b shows an early, strong, and broadly sustained
  curvature arc. In LLaMA, curvature appears early and is followed by
  compact, consistent salience - suggesting fine-grained reorientation
  followed by elaboration. In contrast, Gemma builds momentum through a
  deeper salience wave after a brief early bend. This may reflect a
  simpler representational arc: minimal redirection, followed by
  confident semantic travel.
\item
  \textbf{Semantic geometry reflects both latent potential and
  contextual expression.} We find empirical support for the idea that
  LLM geometry operates across two distinct but interacting levels:

  \begin{itemize}
  \tightlist
  \item
    A \textbf{latent geometry} encoded in the token embedding matrix
    \(E\) and unembedding matrix \(U\) - a static conceptual space
    trained on the corpus.
  \item
    A \textbf{contextual geometry} observed in the residual stream \(x\)
    - the evolving trajectory that expresses meaning in response to
    specific prompts.
  \end{itemize}

  Concern-shifted tokens may often reside in semantically irregular or
  high-potential zones of \(E\) (as shown in recent work by Robinson et
  al.~\href{https://doi.org/10.48550/arXiv.2504.01002}{{[}16{]}}), but
  their actual impact on the model's inference path depends on context.
  Our curvature and salience metrics measure how (and whether) that
  latent potential is activated, resolved, or redirected. This provides
  a natural link between raw embedding structure and downstream semantic
  processing.
\end{enumerate}

\hypertarget{implications-for-interpretability-research}{%
\subsubsection{5.2 Implications for Interpretability
Research}\label{implications-for-interpretability-research}}

These results support a shift toward \textbf{process-level explanations}
of LLM behaviour. Instead of isolated neurons or static feature
attributions, we focus on the model's \textbf{residual stream
trajectory} - tracing how internal representations evolve across layers.
Salience plus curvature and its companions (\textbf{layer-\(\Delta\)}
inter-layer angle change and \textbf{dir-\(\Delta\)} directional
deviation), offer \textbf{continuous, geometry-grounded signals} that
complement discrete methods such as causal tracing or probing.

\textbf{Alignment telemetry.} Because both \(\kappa_i\) and
layer-\(\Delta\) exhibit sharp spikes in response to certain
concernt-shifted prompts, these metrics offer a plausible basis for
\textbf{real-time alignment monitors} - flagging potentially risky
completions \emph{before} they manifest in output.

\textbf{Architecture design.} The contrast between Gemma and LLaMA may
suggest that larger parameter models could defer semantic reorientation
until higher-order representations have stabilised. This would imply
that smaller models may benefit from architectural constraints (such as
delayed residual gating or curvature regularisation) to avoid premature
semantic bending or norm inflation in early layers.

\begin{quote}
\emph{Curvature spikes in some concern-shifted prompts may serve as
real-time alignment signals.}
\end{quote}

\hypertarget{latent-and-contextual-geometry-in-light-of-prior-work}{%
\subsubsection{5.3 Latent and Contextual Geometry in Light of Prior
Work}\label{latent-and-contextual-geometry-in-light-of-prior-work}}

Recent work challenges the assumption that token embeddings lie on a
smooth, low-dimensional manifold. Robinson et
al.~\href{https://doi.org/10.48550/arXiv.2504.01002}{{[}16{]}} present
evidence that LLM embedding spaces violate the \textbf{manifold
hypothesis}, instead exhibiting locally non-Euclidean and singular
behaviour. Their fibre bundle null model formalises how certain tokens
occupy high-noise or irregular subspaces - potentially distorting
semantic inference.

However, in this work they focus on \textbf{static embeddings} - token
vectors from the learned embedding matrix \(E\), unconditioned on prompt
context. In contrast, our \emph{Curved Inference} framework operates in
the \textbf{residual stream}, capturing how meaning is dynamically
expressed across transformer layers. What Robinson labels as
embedding-space singularities may or may not persist during inference.
Some tokens with ``irregular'' positions in \(E\) may be smoothly
integrated through context, while others may spark curvature spikes or
salience surges.

This context-sensitive transformation aligns with findings from vec2vec
\href{https://doi.org/10.48550/arXiv.2505.12540}{{[}17{]}}, which
demonstrates that embeddings from different models can be reliably
aligned via unsupervised translation functions. Despite architectural
differences, a \textbf{shared latent geometry} seems to underlie token
semantics. Vec2vec shows this by learning a mapping from one model's
embedding space to another's, preserving relational structure.

These two findings - \textbf{manifold violation in \(E\)}, and
\textbf{cross-model alignment potential} - are not contradictory.
Instead, they reflect a distinction between \textbf{local irregularity}
and \textbf{global coherence}. \emph{Curved Inference} provides a third,
complementary perspective:

\begin{quote}
\emph{Curved Inference} shows how these latent properties are
\emph{realised} or \emph{suppressed} through context.
\end{quote}

Together, this suggests a layered view of geometry:

\begin{itemize}
\tightlist
\item
  \textbf{Latent conceptual structure} lives in \(E\) and \(U\).
\item
  \textbf{Contextual semantic expression} unfolds in \(x\).
\item
  \textbf{Curved Inference} measures how context bends potential into
  meaning.
\end{itemize}

See section 7 (Future Work) for a discussion of how this perspective
could be explored further.

\hypertarget{limitations}{%
\subsection{6 Limitations}\label{limitations}}

We recognise that several limitations constrain the scope of our
conclusions:

\begin{itemize}
\item
  \textbf{Prompt coverage.}\\
  The concern-shift suite comprises 20 hand-curated prompts across seven
  semantic domains. While balanced for vocabulary and length, the
  dataset remains small and may introduce structural or semantic biases.
  Future work should draw from broader benchmarks (e.g.~MMLU, HELM,
  SuperGLUE) to test generalisability.
\item
  \textbf{Model scale and family.}\\
  This study analyses only two decoder-only checkpoints: Gemma3-1b and
  LLaMA3.2-3b. Larger-capacity models, encoder-decoder hybrids, or
  pretraining variants may bend differently or redistribute curvature
  across attention and MLP submodules. A systematic scaling sweep is
  needed.
\item
  \textbf{Metric sensitivity.}\\
  Curvature and direction-based metrics may respond to \textbf{syntactic
  or lexical changes} that are not genuinely semantic. While residual
  activations show the clearest geometric response, further null
  controls (e.g.~synonym swaps, structure-preserving rewrites) are
  required to confirm the specificity of the signal.
\item
  \textbf{No null baselining.}\\
  While nonsense and neutral prompts were utilised, no scrambled,
  synonym-swapped, and random-weight prompts were tested. Without these
  additional baselines, curvature magnitudes remain \textbf{relatively
  interpreted} - we cannot assign absolute thresholds for ``low'' or
  ``high'' curvature.
\item
  \textbf{Early-layer artefacts.}\\
  Some prompts in Gemma show peak curvature at \textbf{layer 1},
  suggesting possible artefacts from tokenisation, embedding scaling, or
  early-layer normalisation. Without deeper instrumentation, these
  cannot be conclusively labelled as genuine semantic reactions.
  However, it's also possible that concern-sensitive effects may begin
  as early as the embedding layer or at the very point of entry into the
  residual stream. This suggests a mechanism where curvature is seeded
  by prompt-level semantics even prior to token-level integration.
\item
  \textbf{Control prompt assumptions.}\\
  Neutral scaffolds are designed to be semantically flat, but subtle
  phrasal choices may still encode latent salience. This could inflate
  curvature in ``control'' prompts and weaken differential comparisons.
\item
  \textbf{No behavioural validation.}\\
  While curvature patterns are consistent, no task-level metrics were
  collected to confirm causal impact on generation. This limits
  Interpretability claims to internal geometry. Future work could test
  curvature's behavioural relevance via controlled generation studies,
  human evaluation of completions, or causal intervention at
  high-\(\kappa\) points in the residual stream.
\item
  \textbf{Limited modality and feature scope.}\\
  The analysis focuses exclusively on residual stream activations.
  Attention maps, MLP outputs, or intermediate gating signals may also
  express meaningful curvature. Triangulating these modalities could
  refine salience tracking.
\item
  \textbf{Finite-difference sensitivity.} We use a discrete 3-point
  central difference method to estimate curvature. While this is
  effective for short, smooth trajectories, it may become noisy or
  unstable for longer sequences or high-curvature transitions. Using a
  wider finite-difference stencil (e.g.~5-point or 7-point) or
  incorporating smoothing priors could improve robustness on longer
  sequences or noisier trajectories.
\item
  \textbf{Metric scope.}\\
  Our analysis includes extrinsic curvature, layer-wise deviation,
  directional angle, and cosine similarity. Additional geometric
  descriptors (such as \textbf{torsion}, \textbf{energy}, or
  \textbf{intrinsic curvature tensors}) remain unexplored here but offer
  promising extensions.
\item
  \textbf{Finite sequence effects.} Current analysis is limited to short
  prompts. Curvature behaviour may vary with longer sequences. The
  3-point method may also become unstable with longer sequences.
\item
  \textbf{Layer resolution.} With only 26-28 layers, the discrete
  sampling may miss important dynamics. Further experimentation is
  required to validate if this scaled to models with many more layers
  (e.g.~80+).
\end{itemize}

These limitations do not undermine the core findings, but highlight
clear directions for future work: broader prompt and model sampling,
null calibration, multimodal instrumentation, and behavioural grounding
of geometric signals.

\hypertarget{future-work}{%
\subsection{7 Future Work}\label{future-work}}

This study introduces a geometric lens on LLM behaviour, grounded in
path curvature within native activation space. While the framework rests
on principled metric geometry, several promising directions remain:

\begin{itemize}
\item
  \textbf{Singularity and embedding structure.}\\
  The insights from section 5.3 suggest new empirical directions. Tokens
  identified as geometrically singular in \(E\) could be tracked across
  varied prompts to measure how their curvature and salience signatures
  vary by context. Do some contexts neutralise their irregularity? Do
  others amplify it? By combining Robinson's singularity indices with
  our delta metrics, future work could identify not just which tokens
  are charged, but when and how they activate that charge. Such studies
  would further unify pretraining geometry and inference dynamics - and
  clarify how concern becomes curvature.
\item
  \textbf{Interpretive Divergence as Semantic Signal.} One promising
  interpretive direction emerging from the two-layer geometric
  perspective outlined in section 5.3 is the role of
  \textbf{contextual-conceptual divergence} as a marker of semantic
  significance. While this paper focuses on how concern shifts alter the
  shape of residual trajectories, future work could explore when and why
  a token's \textbf{contextual meaning trajectory diverges sharply} from
  its \textbf{latent conceptual position}. Such divergence may signal
  both \textbf{positive behaviours} (e.g.~abstraction, creativity, novel
  synthesis) and \textbf{failure modes} (e.g.~hallucination,
  incoherence, misalignment). Tokens used in ways that sharply depart
  from their pretraining priors may produce high curvature relative to
  their embedding-based expectation. By measuring this
  conceptual-contextual gap (using angular divergence, curvature, or
  alignment metrics) \emph{Curved Inference} could support new forms of
  semantic anomaly detection and creativity tracing.
\item
  \textbf{Force attribution.} While Appendix A frames attention and MLP
  block layers as ``semantic lenses'' that induce the observed curvature
  in the residual stream, we leave a detailed force-alignment
  study-e.g., correlating per-layer MLP output with discrete
  accelerations, or ablating high-impact attention heads-to future work.
\item
  \textbf{Causal interventions.}\\
  Whether curvature reflects a causal locus of computation remains
  unclear. Having identified high-curvature layers aligned with concern
  tokens, future work could test their functional role through patching,
  ablation, or targeted editing.
\item
  \textbf{Scaling and architectural generalisation.}\\
  This study focused on two decoder-only models (1b-3b). Extending the
  analysis to larger checkpoints and different model families would test
  the generality of observed curvature behaviours.
\item
  \textbf{Intrinsic geometric structure.}\\
  While this work focused on extrinsic curvature in full-dimensional
  space, it's possible that token trajectories lie near a
  lower-dimensional manifold. Exploring intrinsic curvature or geodesic
  deviation could uncover deeper representational geometry.
\item
  \textbf{Alignment and robustness.}\\
  Because curvature often spikes on morally or emotionally charged
  prompts, it may correlate with semantic load or alignment risk.
  Applying curvature analysis to robustness benchmarks or generation
  pipelines could test its use as a real-time safety signal.
\item
  \textbf{Cross-lingual and cross-domain generalisation.}\\
  Curvature signals should be tested across languages, tokenisation
  regimes, and domains. Prior work shows residual trajectories align
  across translations \href{https://arxiv.org/abs/2405.15943}{{[}18{]}};
  future work could test whether curvature shape generalises similarly,
  revealing a universal form of concern.
\end{itemize}

Together, these directions will help test whether curvature is not just
diagnostic - but generative, transferable, and actionable.

\hypertarget{conclusions}{%
\subsection{8 Conclusions}\label{conclusions}}

This work introduces \emph{Curved Inference} - a geometry-first
framework for analysing how large language models bend their internal
representational space in response to latent \emph{concern}. By
grounding all measurements in native residual space \(\mathbb{R}^d\) and
computing path curvature \(\kappa_i\) and salience \(S(t)\) using the
semantic pullback metric \(G = U^\top U\), we eliminate potential
distortions introduced by low-dimensional projections and recover
meaningful geometric structure.

Our empirical results confirm that concern-shifted prompts induce
measurable and model-specific changes in the internal trajectories of
LLMs. In particular:

\begin{itemize}
\tightlist
\item
  \textbf{LLaMA3.2-3b} displays clear, statistically significant scaling
  of curvature and salience in response to increasing concern strength.
\item
  \textbf{Gemma3-1b}, while reactive to concern shifts, shows weaker
  differentiation between moderate and strong variants.
\end{itemize}

These patterns reflect deeper architectural and geometric dynamics,
captured through our proposed distinction between two interacting layers
of semantic geometry:

\begin{itemize}
\tightlist
\item
  A \textbf{latent conceptual geometry}, encoded in token embeddings
  (\(E\)) and unembedding projections (\(U\)), shaped during
  pretraining.
\item
  A \textbf{contextual semantic geometry}, expressed through residual
  trajectories (\(x\)) during inference.
\end{itemize}

We show that while some tokens occupy irregular or singular regions in
embedding space (as described by Robinson et
al.~\href{https://doi.org/10.48550/arXiv.2504.01002}{{[}16{]}}), their
impact is determined by how context interacts with that potential.
Meanwhile, the vec2vec
\href{https://doi.org/10.48550/arXiv.2505.12540}{{[}17{]}} findings
reveal that despite architectural divergence, different models often
encode semantically compatible latent structures - a finding that our
residual-path analysis helps operationalise.

\emph{Curved Inference} provides a principled method for tracing how
\textbf{semantic potential becomes semantic movement}. Its native-space
metrics expose how meaning is navigated, reoriented, or reinforced
within a model's depth. These geometric insights complement traditional
Interpretability methods, and open new paths for alignment monitoring,
architectural diagnosis, and semantic probing.

Earlier versions of this work used UMAP-based projections to visualise
geometric divergence. These proved inadequate, mischaracterising the
timing and scale of curvature effects. This led to the
mischaracterisation of LLaMA's curvature as ``late and steep''.
Full-space analysis using \(\kappa_i\) and \(S(t)\) corrected this,
revealing that curvature is strongest \textbf{early} in the residual
trajectory and often sustained across depth - a correction that
highlights the importance of measuring semantic geometry in native
activation space.

In mapping internal semantic geometry, \emph{Curved Inference} shifts
Interpretability away from static probes and component dissection and
toward \textbf{dynamic trajectory analysis} - a vantage point that
scales with model size and abstraction depth.

\emph{Curved Inference} is both a \textbf{map} and a
\textbf{diagnostic}:

\begin{quote}
It reveals what the model finds meaningful - and how it bends to meet
it.
\end{quote}

\hfill\break
\hfill\break
\hfill\break
\hfill\break
\hfill\break
\hfill\break
\hfill\break
\hfill\break
\hfill\break

\hypertarget{references}{%
\subsection{References}\label{references}}

1 - \href{https://doi.org/10.48550/arXiv.2312.12141}{\textbf{Yu, Z., et
al.} (2023) ``Exploring the Residual Stream of Transformers''
\emph{arXiv}}

2 - \href{https://doi.org/10.48550/arXiv.1902.10186}{\textbf{Jain, S. \&
Wallace, B.} (2019) ``Attention is not Explanation'' \emph{arXiv}}

3 - \href{https://doi.org/10.48550/arXiv.1805.01070}{\textbf{Conneau,
A., et al.} (2018) ``What you can cram into a single vector: Probing
sentence embeddings for linguistic properties'' \emph{arXiv}}

4 - \href{https://distill.pub/2020/circuits/zoom-in/}{\textbf{Olah, C.,
et al.} (2020) ``Zoom In: An Introduction to Circuits'' \emph{Distill}}

5 -
\href{https://transformer-circuits.pub/2021/framework/index.html}{\textbf{Elhage,
N., et al.} (2021) ``A Mathematical Framework for Transformer Circuits''
\emph{arXiv}}

6 - \href{https://doi.org/10.48550/arXiv.2405.05386}{\textbf{Madsen, A.
et al.} (2024) ``Interpretability Needs a New Paradigm'' \emph{arXiv}}

7 - \href{https://doi.org/10.48550/arXiv.2402.01761}{\textbf{Singh, C.
et al.} (2024) ``Rethinking Interpretability in the Era of Large
Language Models'' \emph{arXiv}}

8 - \href{https://doi.org/10.48550/arXiv.2402.11863}{\textbf{Yeo, W. et
al.} (2024) ``How Interpretable are Reasoning Explanations from
Prompting Large Language Models?'' \emph{arXiv}}

9 - \href{https://doi.org/10.48550/arXiv.2310.11207}{\textbf{Huang, S.
et al.} (2023) ``Can Large Language Models Explain Themselves? A Study
of LLM-Generated Self-Explanations'' \emph{arXiv}}

10 - \href{https://doi.org/10.48550/arXiv.2309.07315}{\textbf{Molina,
R.} (2023) ``Traveling Words: A Geometric Interpretation of
Transformers'' \emph{arXiv}}

11 - \href{https://doi.org/10.48550/arXiv.2405.15943}{\textbf{Shai, A.
et al.} (2025) ``Transformers Represent Belief State Geometry in their
Residual Stream'' \emph{arXiv}}

12 - \href{https://github.com/robman/FRESH-model}{\textbf{Manson, R.}
(2025) ``FRESH (Functionalist \& Representationalist Emergent Self
Hypothesis)'' \emph{Github}}

13 -
\href{https://github.com/robman/FRESH-model/blob/main/benchmarks/curved-inference/01/}{\textbf{Manson,
R.} (2025) ``\emph{Curved Inference} in LLMs - Experiment''
\emph{Github}}

14 - \href{https://ai.google.dev/gemma/docs/core}{\textbf{Google} (2025)
``Gemma 3 Model Card'' \emph{ai.google.dev}}

15 -
\href{https://github.com/meta-llama/llama-models/blob/main/models/llama3_2/MODEL_CARD.md}{\textbf{Meta}
(2024) ``LLaMA 3.2 Model Card'' \emph{Github}}

16 - \href{https://doi.org/10.48550/arXiv.2504.01002}{\textbf{Robinson,
M., et al.} (2025) ``Token Embeddings Violate the Manifold Hypothesis''
\emph{arXiv}}

17 - \href{https://doi.org/10.48550/arXiv.2505.12540}{\textbf{Jha, R.,
et al.} (2025) ``Harnessing the Universal Geometry of Embeddings''
*arXiv}

18 - \href{https://arxiv.org/abs/2405.15943}{\textbf{Tian, Y., et al.}
(2024) ``Neural Interlinguae: Layerwise Geometry Aligns Translation
Trajectories Across Languages'' \emph{arXiv}}

19 -
\href{https://transformer-circuits.pub/2023/privileged-basis/index.html}{\textbf{Elhage,
N., et al.} (2023) ``A Privileged Basis for the Transformer Residual
Stream'' \emph{Transformer Circuits}}

\hfill\break
\hfill\break
\hfill\break
\hfill\break
\hfill\break
\hfill\break
\hfill\break
\hfill\break
\hfill\break
\hfill\break
\hfill\break
\hfill\break
\hfill\break
\hfill\break
\hfill\break
\hfill\break
\hfill\break
\hfill\break
\hfill\break
\hfill\break
\hfill\break

\hypertarget{appendix-a---semantic-geometry-of-transformer-inference}{%
\subsection{\texorpdfstring{\textbf{Appendix A} - Semantic Geometry of
Transformer
Inference}{Appendix A - Semantic Geometry of Transformer Inference}}\label{appendix-a---semantic-geometry-of-transformer-inference}}

\hypertarget{a.1-overview-geometry-as-trajectory}{%
\subsubsection{A.1 Overview: Geometry as
Trajectory}\label{a.1-overview-geometry-as-trajectory}}

Transformer inference can be viewed as a geometric process - each token
is mapped to a vector in a high-dimensional space, and then pushed
through a series of attention and MLP updates. The result is a
continuous sequence of transformations forming a trajectory in residual
space. This trajectory encodes the evolving semantic state of the model
as it processes or generates a sequence. Attention and MLP layers act as
dynamic lenses, bending and focusing these trajectories based on
contextual and relational signals. This appendix outlines how these
trajectories are constructed, how they evolve, and how geometric
measurements such as curvature and salience are defined within this
process.

\begin{quote}
\textbf{Notation Recap:}

\begin{itemize}
\tightlist
\item
  \(E\): embedding matrix, maps token IDs to vectors in \(\mathbb{R}^d\)
\item
  \(U\): unembedding matrix, maps residual vectors to logit space (often
  \(U = E\))
\item
  \(x\): residual stream vector, the current semantic state
\item
  \(l = Ux\): logit vector (dot products between \(x\) and each token
  direction)
\item
  \(G = U^\top U\): pullback metric from logit space, defines geometry
  in residual space
\end{itemize}
\end{quote}

\hypertarget{a.2-token-trajectories-and-residual-flow}{%
\subsubsection{A.2 Token Trajectories and Residual
Flow}\label{a.2-token-trajectories-and-residual-flow}}

Tokens are first split by a tokeniser and mapped to unique token IDs.
Each token ID is used to look up an embedding vector from the learned
embedding matrix \(E \in \mathbb{R}^{|V| \times d}\), where \(|V|\) is
the vocabulary size and \(d\) is the model dimension.

The initial residual stream vector \(x \in \mathbb{R}^d\) for a token is
simply the embedding \(E[t]\). In models using Rotary Positional
Embedding (RoPE), no position vector is added at this stage. Instead,
positional information is injected later during attention via rotation.

At each transformer layer, the residual vector is updated by adding the
outputs of the attention and MLP sublayers:

\[
x^{(\ell+1)} = x^{(\ell)} + \text{Attention}(x^{(\ell)}) + \text{MLP}(x^{(\ell)})
\]

This additive structure means that the residual stream forms a
trajectory through \(\mathbb{R}^d\), with each step determined by the
semantic influence of the attention and MLP mechanisms.

\begin{itemize}
\tightlist
\item
  The attention layer gathers contextual signals from other positions,
  modulated by relative position (via RoPE), and contributes a vector
  update that reflects token-token interaction.
\item
  The MLP layer applies a local, nonlinear transformation that sharpens
  or redirects the vector - often enhancing its alignment with
  task-relevant directions in the model.
\end{itemize}

Together, these updates shape the path taken by each token's
representation. We refer to this evolving path as the token's semantic
trajectory.

Because each update is added to the previous residual state, it is only
in the residual stream that one can observe the full evolution of
meaning over depth. Attention and MLP outputs are delta vectors - they
cause curvature, but the residual stream \emph{is} the curve.

\hypertarget{a.3-position-rope-and-curvature}{%
\subsubsection{A.3 Position, RoPE, and
Curvature}\label{a.3-position-rope-and-curvature}}

In models using RoPE, positional information is not embedded additively
into the residual vector. Instead, RoPE applies a deterministic,
sinusoidal rotation to the query and key vectors used in attention.
These rotations encode relative position by angular offset, preserving
dot products while modulating attention scores.

Because RoPE does not shift or perturb the initial residual vector, it
preserves semantic purity in the early layers. Curvature in the residual
stream only arises once RoPE-modulated attention begins to redistribute
contextual information across tokens.

This means that curvature is not tied to absolute position, but to
\emph{semantic interaction} among tokens that are contextually relevant
and positionally adjacent. As a result, the residual trajectory bends
not because of where a token is, but because of how it relates to
others. RoPE thereby becomes a key generator of semantically aligned
curvature.

\hypertarget{a.4-logits-and-the-pullback-metric}{%
\subsubsection{A.4 Logits and the Pullback
Metric}\label{a.4-logits-and-the-pullback-metric}}

Once the final residual vector is computed for a token, it is projected
into logit space by taking a dot product with each row of the
unembedding matrix \(U \in \mathbb{R}^{|V| \times d}\):

\[
l = Ux
\]

This yields a logit vector \(l \in \mathbb{R}^{|V|}\), where each entry
reflects the alignment between \(x\) and a possible output token
direction \(u_i\). The softmax of \(l\) forms a probability distribution
over the vocabulary.

Crucially, the matrix \(U\) defines a set of semantic directions in
residual space. These directions induce a geometry - the pullback metric
\(G = U^\top U\) redefines how distances and angles are measured in
residual space based on the model's output behaviour.

This metric allows curvature to be computed in a way that reflects
semantic change. Rather than relying on arbitrary coordinates, we
measure changes in the residual trajectory using a geometry aligned with
token prediction. This is what gives curvature its interpretability - it
reflects changes in internal intent as judged by the model's own output
semantics.

\hypertarget{a.5-caching-and-contextual-reuse}{%
\subsubsection{A.5 Caching and Contextual
Reuse}\label{a.5-caching-and-contextual-reuse}}

During inference, transformer models use key-value (KV) caching to avoid
recomputing attention outputs for previously processed tokens. Once a
token's attention keys and values have been computed, they are stored
and reused in subsequent steps. This optimisation ensures that only the
\emph{new} token's computations need to be performed at each generation
step.

Importantly, this also means that all previous residual vectors are
frozen - they are not recomputed or updated. Each prior \(x\) forms a
fixed semantic anchor. The residual stream for the current token builds
on top of these fixed vectors, enabling us to track how each new token
evolves in context.

In multi-turn chat settings, the entire chat history is tokenised into a
flat prompt. Provided it fits within the context window, only the new
portion of the prompt (e.g.~user query and assistant response) is
recomputed. The rest is reused, including residuals, keys, and values.

\hypertarget{a.6-summary-the-semantic-lens}{%
\subsubsection{A.6 Summary: The Semantic
Lens}\label{a.6-summary-the-semantic-lens}}

The transformer can be viewed as a geometric engine. Tokens enter as
points in a semantic subspace, pass through layers of contextual and
nonlinear modulation, and exit as probability distributions over token
space. The residual stream traces the continuous trajectory of each
token through this process.

Attention and MLP layers act as semantic lenses. Attention bends
trajectories based on relative semantic and positional relevance. MLPs
sharpen or redirect them through nonlinear amplification. RoPE enables
these transformations to be position-aware without distorting the
embedding space directly.

All curvature, salience, and concern arise within the residual stream.
It is the only continuous representational path through the model-and
the only space in which geometric measurements can meaningfully be made.

\hfill\break
\hfill\break
\hfill\break
\hfill\break
\hfill\break
\hfill\break
\hfill\break
\hfill\break

\hypertarget{appendix-b---definitions}{%
\subsection{\texorpdfstring{\textbf{Appendix B} -
Definitions}{Appendix B - Definitions}}\label{appendix-b---definitions}}

\hypertarget{meaning}{%
\subsubsection{Meaning}\label{meaning}}

\textbf{Meaning}, in the context of large language models, refers to the
implicit content, intent, or conceptual structure represented by the
model's internal activations. It is not a directly observable quantity,
but an abstract property inferred from the model's behaviour and
internal geometry.

\hypertarget{conceptual-role}{%
\paragraph{\texorpdfstring{Conceptual Role\\
}{Conceptual Role }}\label{conceptual-role}}

\hfill\break
Meaning is what the model \emph{represents} at any point in the forward
pass. This could include factual information, sentiment, identity,
logical structure, or moral stance. The meaning associated with a given
activation depends on the context, the model's training, and how that
activation aligns with downstream predictions.

Meaning becomes accessible through \textbf{semantic structure} - the way
that internal representations relate to each other and to output tokens.
This structure is revealed through geometric properties-such as
direction, distance, and curvature-within the residual stream.

\hypertarget{formal-proxy}{%
\paragraph{\texorpdfstring{Formal Proxy\\
}{Formal Proxy }}\label{formal-proxy}}

\hfill\break
We do not measure meaning directly. Instead, we study how it
\textbf{moves} and \textbf{changes} through time. This is done by:

\begin{itemize}
\item
  Representing the model's internal state as a vector
  \(x_t \in \mathbb{R}^d\)
\item
  Measuring change (salience) as \(\|x_{t+1} - x_t\|_G\)
\item
  Measuring reorientation (curvature) as \(\kappa_i\)
\item
  Anchoring this geometry in \textbf{semantic space} via the pullback
  metric:

  \[
  G = U^\top U
  \]
\end{itemize}

This metric ensures that the geometry of residual-space movement
reflects differences in token-level output probabilities. In this sense,
\textbf{meaning lives in the structure} of how internal representations
flow and bend toward predicted outputs.

\hypertarget{practical-implication}{%
\paragraph{\texorpdfstring{Practical Implication\\
}{Practical Implication }}\label{practical-implication}}

\hfill\break
Throughout this work, we treat \textbf{meaning} as:

\begin{quote}
\emph{The internal state of the model that gives rise to token
predictions and reflects the model's interpretation of context.}
\end{quote}

Changes in meaning are inferred from changes in the residual trajectory.
High salience means meaning is shifting quickly; high curvature means it
is changing direction. Concern identifies directions along which meaning
changes matter to the model.

This view of meaning intersects with the idea of superposition - that
many abstract features may be simultaneously encoded in overlapping
directions within the same residual vector. The geometric structure
(e.g.~curvature) reflects how these meanings are separated or recombined
across layers.

\hypertarget{semantic-space}{%
\subsubsection{Semantic Space}\label{semantic-space}}

\textbf{Semantic space} refers to the internal vector space in which a
model encodes and manipulates meaning. It is the geometric arena where
representations of language, context, and concepts take shape and evolve
during inference. In transformer-based LLMs, this space is typically
identified with the \textbf{residual stream} - but only when measured
under a \textbf{meaning-preserving metric}.

\hypertarget{formal-definition}{%
\paragraph{\texorpdfstring{Formal Definition\\
}{Formal Definition }}\label{formal-definition}}

\hfill\break
We define semantic space as the residual space \(\mathbb{R}^d\),
equipped with a metric derived from the model's output behaviour:

\[
G = U^\top U
\]

where \(U \in \mathbb{R}^{V \times d}\) is the unembedding matrix that
projects residual activations \(x_t \in \mathbb{R}^d\) to logits over
the vocabulary. The inner product and norm induced by \(G\) give rise to
a geometry in which:

\begin{itemize}
\tightlist
\item
  \textbf{Distances} correspond to shifts in output token probabilities
\item
  \textbf{Directions} correspond to latent semantic operations
\item
  \textbf{Curves} correspond to evolving meaning across layers
\end{itemize}

This pullback metric transforms residual space into a
\textbf{logit-aligned semantic space}.

\hypertarget{distinctions-and-relationships}{%
\paragraph{\texorpdfstring{Distinctions and Relationships\\
}{Distinctions and Relationships }}\label{distinctions-and-relationships}}

\hfill\break
\textbf{Table 6 Comparison of Spaces}

\begin{longtable}[]{@{}
  >{\raggedright\arraybackslash}p{(\columnwidth - 4\tabcolsep) * \real{0.1765}}
  >{\raggedright\arraybackslash}p{(\columnwidth - 4\tabcolsep) * \real{0.3529}}
  >{\raggedright\arraybackslash}p{(\columnwidth - 4\tabcolsep) * \real{0.4706}}@{}}
\toprule\noalign{}
\begin{minipage}[b]{\linewidth}\raggedright
Space
\end{minipage} & \begin{minipage}[b]{\linewidth}\raggedright
Contents
\end{minipage} & \begin{minipage}[b]{\linewidth}\raggedright
Function
\end{minipage} \\
\midrule\noalign{}
\endhead
\bottomrule\noalign{}
\endlastfoot
Embedding space & Token embeddings \(e_i\) & Stores static lexical
representations \\
Logit space & Output predictions \(\ell = Ux\) & Determines token-level
output \\
Residual space & Internal state \(x_t\) & Active inference trajectory \\
Semantic space & Residual space + \(G\) metric & Geometry aligned with
meaning and output \\
\end{longtable}

Semantic space is not defined purely by coordinate axes-it emerges from
the \textbf{functional role} of directions and distances under the
model's output logic. That is, it reflects how the model internally
represents and differentiates concepts, rather than any superficial
arrangement of neurons.

\hypertarget{practical-use-in-this-work}{%
\paragraph{\texorpdfstring{Practical Use in This Work\\
}{Practical Use in This Work }}\label{practical-use-in-this-work}}

\hfill\break
All geometric quantities in this paper (salience, curvature, directional
shifts), are computed in \textbf{semantic space}, using the pullback
metric \(G\). This ensures that our analysis reflects what the model
\emph{does} with its internal states, not just how they appear
numerically.

\hypertarget{concern}{%
\subsubsection{Concern}\label{concern}}

In this paper, we study concern \textbf{extrinsically} by introducing
controlled shifts in prompt semantics - what we call
\textbf{concern-shifted prompts}. These allow us to probe whether and
how the model reorients its internal trajectory in response to targeted
emotional, moral, or logical pressure. While this is an external
perturbation, we interpret the resulting geometric responses as signals
of \textbf{internal concern sensitivity}.

\hypertarget{operational-definition-this-paper}{%
\paragraph{\texorpdfstring{Operational Definition (This Paper)\\
}{Operational Definition (This Paper) }}\label{operational-definition-this-paper}}

\hfill\break
In this study, concern is defined \textbf{extrinsically} through a
controlled prompt-design strategy. Each scaffold contains a
\textbf{neutral base} plus \textbf{concern-shifted prompts}, chosen to
introduce a targeted form of concern (e.g.~moral framing, emotional
tone, identity cue). The resulting \textbf{geometric divergence}
(measured via residual stream metrics such as path curvature
\(\kappa_i\)), is interpreted as a behavioural signal of the model's
sensitivity to that shift.

This approach allows us to identify \textbf{concern-induced curvature},
where the model's internal trajectory bends or accelerates in response
to semantic pressure, even if token-level differences are minimal.

\hypertarget{conceptual-distinction}{%
\paragraph{\texorpdfstring{Conceptual Distinction\\
}{Conceptual Distinction }}\label{conceptual-distinction}}

\hfill\break
Concern is orthogonal to salience and curvature.

\textbf{Table 7 Comparison of 3 Primary Concepts}

\begin{longtable}[]{@{}
  >{\raggedright\arraybackslash}p{(\columnwidth - 6\tabcolsep) * \real{0.1163}}
  >{\raggedright\arraybackslash}p{(\columnwidth - 6\tabcolsep) * \real{0.2558}}
  >{\raggedright\arraybackslash}p{(\columnwidth - 6\tabcolsep) * \real{0.3643}}
  >{\raggedright\arraybackslash}p{(\columnwidth - 6\tabcolsep) * \real{0.2636}}@{}}
\toprule\noalign{}
\begin{minipage}[b]{\linewidth}\raggedright
Term
\end{minipage} & \begin{minipage}[b]{\linewidth}\raggedright
Type
\end{minipage} & \begin{minipage}[b]{\linewidth}\raggedright
What it captures
\end{minipage} & \begin{minipage}[b]{\linewidth}\raggedright
Example
\end{minipage} \\
\midrule\noalign{}
\endhead
\bottomrule\noalign{}
\endlastfoot
Salience & First-order (\(\|x_{t+1} - x_t\|\)) & How fast the model's
state is changing & Rapid progression through a narrative \\
Curvature & Second-order (\(\kappa_i\)) & Whether the trajectory is
turning & Reorienting after a moral twist \\
Concern & Priority weighting & Whether the model treats this direction
as significant & Responding to identity or moral cues \\
\end{longtable}

\hypertarget{potential-future-intrinsic-formulations}{%
\paragraph{\texorpdfstring{Potential Future (Intrinsic) Formulations\\
}{Potential Future (Intrinsic) Formulations }}\label{potential-future-intrinsic-formulations}}

\hfill\break
Longer-term, we aim to explore replacing the extrinsic scaffold design
with \textbf{intrinsic concern measures} derived from model-internal
behaviour:

\begin{itemize}
\item
  \textbf{Gradient-based sensitivity}:

  \[
  \text{Concern}(x_t) = \left\| \frac{\partial \ell}{\partial x_t} \right\|_G
  \]

  Captures how sensitive the output is to changes at layer \(t\),
  measured in the pullback metric.
\item
  \textbf{Subspace projection}:

  \[
  \text{Concern}(x_t) = \|P_{\mathcal{C}} x_t\|_G
  \]

  Where \(\mathcal{C}\) is a concern-relevant subspace extracted via
  PCA, CCA, or probing.
\item
  \textbf{Behavioural perturbation}: KL divergence between softmax
  outputs after small displacements in specific directions.
\end{itemize}

Each of these aims to isolate \emph{what the model finds meaningful}, as
inferred from its own structure and behaviour.

\hypertarget{salience}{%
\subsubsection{Salience}\label{salience}}

\textbf{Salience} quantifies how rapidly a model's internal state is
changing as it processes a prompt. In geometric terms, it is the
\textbf{first-order velocity} of the residual stream trajectory - how
far the model moves in semantic space from one layer to the next. High
salience indicates a rapid update in the model's internal
representation, even if that movement follows a straight path.

\hypertarget{operational-definition}{%
\paragraph{\texorpdfstring{Operational Definition\\
}{Operational Definition }}\label{operational-definition}}

\hfill\break
For a residual stream trajectory
\({x_0, x_1, \dots, x_L} \subset \mathbb{R}^d\), the \textbf{layer-wise
salience} at layer \(t\) is defined as:

\[
\text{Salience}(t) = \|x_{t+1} - x_t\|_G
\]

where \(\| \cdot \|_G\) denotes the norm induced by the \textbf{semantic
metric}:

\[
G = U^\top U
\]

Here, \(U\) is the unembedding matrix that maps residual states to
logits, and the pullback metric \(G\) aligns geometric measurements in
residual space with token-level semantic structure.

\hypertarget{semantic-interpretation}{%
\paragraph{\texorpdfstring{Semantic Interpretation\\
}{Semantic Interpretation }}\label{semantic-interpretation}}

\hfill\break
Salience tracks the \textbf{rate of change of internal meaning}, where
``meaning'' is defined by how the residual vector projects into logit
space. It captures \emph{how much} the model updates its belief or
understanding at each layer - irrespective of direction.

A model may have:

\begin{itemize}
\tightlist
\item
  \textbf{High salience, low curvature} → confidently elaborating or
  reinforcing an idea
\item
  \textbf{Low salience, high curvature} → making a subtle but meaningful
  reorientation
\item
  \textbf{Low salience, low curvature} → continuing steadily with no
  shift in interpretation
\end{itemize}

\hypertarget{aggregation}{%
\paragraph{\texorpdfstring{Aggregation\\
}{Aggregation }}\label{aggregation}}

\hfill\break
The total salience over a trajectory can be defined as cumulative arc
length:

\[
S = \sum_{t=0}^{L-1} \|x_{t+1} - x_t\|_G
\]

This is used in later analysis (e.g., for arc-length normalisation in
curvature metrics).

\hypertarget{curvature}{%
\subsubsection{Curvature}\label{curvature}}

\textbf{Curvature} captures how sharply the model's internal
representation is \textbf{changing direction} as it processes a prompt.
In geometric terms, it is the \textbf{second-order property} of the
residual stream trajectory - the rate at which the model's semantic path
bends, rather than continues in a straight line.

\hypertarget{operational-definition-1}{%
\paragraph{\texorpdfstring{Operational Definition\\
}{Operational Definition }}\label{operational-definition-1}}

\hfill\break
Let the residual stream activations across layers be denoted:

\[
{x_0, x_1, \dots, x_L} \subset \mathbb{R}^d
\]

To estimate curvature, we apply a discrete 3-point finite-difference
scheme to the sequence of residual stream vectors. For each interior
point \(i\), we compute the first and second derivatives using a
discrete 3-point central difference method that accounts for unequal
step sizes, then apply the standard extrinsic curvature formula.

The \textbf{extrinsic curvature} at index \(i\) is defined as:

\[
\kappa_i =
\frac{
\sqrt{
\|a_i\|_G^2 \cdot \|v_i\|_G^2 - \langle a_i, v_i \rangle_G^2
}
}{
\|v_i\|_G^3
}
\]

where:

\begin{itemize}
\tightlist
\item
  \(v_i\) is the first derivative (velocity)
\item
  \(a_i\) is the second derivative (acceleration)
\item
  \(\langle \cdot, \cdot \rangle_G\) and \(| \cdot |_G\) denote the
  inner product and norm under the pullback metric \(G = U^\top U\)
\end{itemize}

This is the standard formula for curvature in Euclidean space, extended
here to a semantically aligned geometry via the metric \(G\). It is
invariant to orthogonal coordinate transformations and reflects
intrinsic trajectory shape rather than coordinate artefacts.

\hypertarget{semantic-interpretation-1}{%
\paragraph{\texorpdfstring{Semantic Interpretation\\
}{Semantic Interpretation }}\label{semantic-interpretation-1}}

\hfill\break
Whereas \textbf{salience} measures how far the model moves between
steps, \textbf{curvature} measures how much it \textbf{reorients}
e.g.~whether the model continues in a consistent direction or turns
sharply at some layer. High curvature indicates a structural shift in
internal representation, such as a reinterpretation, contradiction, or
redirection in meaning.

Examples:

\begin{itemize}
\tightlist
\item
  A strong moral reversal → high curvature
\item
  Steady elaboration of a factual detail → low curvature
\end{itemize}

\hypertarget{aggregation-and-summary-statistics}{%
\paragraph{\texorpdfstring{Aggregation and Summary Statistics\\
}{Aggregation and Summary Statistics }}\label{aggregation-and-summary-statistics}}

\hfill\break
From the full curvature series \(\kappa_i\), we derive summary metrics
per prompt variant:

\begin{itemize}
\item
  \textbf{Mean curvature}:

  \[
  \bar{\kappa} = \frac{1}{L - 1} \sum_{i=1}^{L - 1} \kappa_i
  \]
\item
  \textbf{Maximum curvature}:

  \[
  \kappa_{\text{max}} = \max_i \kappa_i
  \]
\item
  \textbf{Layer of maximum curvature}:

  \[
  i^* = \arg\max_i \kappa_i
  \]
\end{itemize}

Curvature is only defined at interior indices \(i\), where a discrete
3-point central difference can be used to estimate derivatives. Boundary
positions \(i = 0\) and \(i = L\) are excluded because symmetric
differencing is not possible.

\hypertarget{comparison}{%
\subsubsection{Comparison}\label{comparison}}

The following table summarises the key interpretive and mathematical
roles of the five core terms used throughout this paper.

\textbf{Table 8 Comparison of Interpretive and Mathematical Roles of Key
Terms}

\begin{longtable}[]{@{}
  >{\raggedright\arraybackslash}p{(\columnwidth - 6\tabcolsep) * \real{0.1724}}
  >{\raggedright\arraybackslash}p{(\columnwidth - 6\tabcolsep) * \real{0.1552}}
  >{\raggedright\arraybackslash}p{(\columnwidth - 6\tabcolsep) * \real{0.3218}}
  >{\raggedright\arraybackslash}p{(\columnwidth - 6\tabcolsep) * \real{0.3506}}@{}}
\toprule\noalign{}
\begin{minipage}[b]{\linewidth}\raggedright
Term
\end{minipage} & \begin{minipage}[b]{\linewidth}\raggedright
Type / Role
\end{minipage} & \begin{minipage}[b]{\linewidth}\raggedright
What It Captures
\end{minipage} & \begin{minipage}[b]{\linewidth}\raggedright
Operational Form
\end{minipage} \\
\midrule\noalign{}
\endhead
\bottomrule\noalign{}
\endlastfoot
\textbf{Meaning} & Latent content & The internal representation of
context, concepts, and intent & Inferred from geometry and output
alignment \\
\textbf{Semantic space} & Geometric setting & The metric space in which
meaning is encoded and compared & Residual space with pullback metric
\(G = U^\top U\) \\
\textbf{Salience} & First-order (velocity) & How rapidly the model's
internal state is changing & \(\|x_{t+1} - x_t\|_G\) \\
\textbf{Curvature} & Second-order (reorientation) & How sharply the
model changes direction in semantic space & \(\kappa_i\) from discrete
3-point finite differences \\
\textbf{Concern} & Directional importance & Whether a direction is
semantically or behaviourally significant & Currently via prompt-class
variation - future via gradients or projections \\
\end{longtable}

This taxonomy supports a structured analysis of model behaviour in
geometric terms - \emph{salience} describes motion, \emph{curvature}
describes trajectory shape, and \emph{concern} identifies which
directions matter. All are grounded in \emph{semantic space}, which
gives these geometric properties functional meaning in terms of the
model's outputs.

\hypertarget{supporting-definitions}{%
\subsubsection{Supporting Definitions}\label{supporting-definitions}}

\hypertarget{semantic-metric}{%
\paragraph{\texorpdfstring{Semantic Metric\\
}{Semantic Metric }}\label{semantic-metric}}

\hfill\break
We define the pullback metric \(( G = U^\top U )\) on residual space,
where \(( U )\) is the unembedding matrix from residual space to logits.
This equips residual activations with a geometry that respects the
model's token-level semantics. All distances, angles, and curvatures are
computed under this metric.

This pullback metric approach reflects a broader insight that residual
directions are not functionally uniform - some carry disproportionate
semantic weight as shown in Elhage et
al.~\href{https://transformer-circuits.pub/2023/privileged-basis/index.html}{{[}19{]}}
through their analysis of privileged basis vectors.

\hypertarget{notation}{%
\paragraph{\texorpdfstring{Notation\\
}{Notation }}\label{notation}}

\begin{itemize}
\tightlist
\item
  \((x_t \in \mathbb{R}^d)\): residual stream activation at layer \(t\)
\item
  \((x_i)\): residual stream activation at layer \(i\), forming a
  discrete trajectory in \(\mathbb{R}^d\)
\item
  \((\kappa_i)\): discrete curvature at interior index \(i\), computed
  from local 3-point finite differences
\item
  \((v_i)\): estimated velocity at index \(i\) (first derivative of
  \(x\))
\item
  \((a_i)\): estimated acceleration at index \(i\) (second derivative of
  \(x\))
\item
  \((G = U^\top U)\): semantic (pullback) metric
\end{itemize}

\hfill\break
\hfill\break
\hfill\break
\hfill\break
\hfill\break
\hfill\break
\hfill\break
\hfill\break
\hfill\break
\hfill\break
\hfill\break
\hfill\break
\hfill\break
\hfill\break
\hfill\break
\hfill\break
\hfill\break
\hfill\break
\hfill\break
\hfill\break
\hfill\break
\hfill\break
\hfill\break
\hfill\break
\hfill\break
\hfill\break
\hfill\break
\hfill\break
\hfill\break
\hfill\break
\hfill\break
\hfill\break
\hfill\break

\hypertarget{appendix-c---discrete-curvature-geometric-criterion-for-semantic-divergence}{%
\subsection{\texorpdfstring{\textbf{Appendix C} - Discrete Curvature:
Geometric Criterion for Semantic
Divergence}{Appendix C - Discrete Curvature: Geometric Criterion for Semantic Divergence}}\label{appendix-c---discrete-curvature-geometric-criterion-for-semantic-divergence}}

This appendix provides a minimal geometric condition under which a
concern-shifted prompt must exhibit curvature in the model's residual
stream trajectory.

\hypertarget{c.1-core-result}{%
\subsubsection{C.1 Core Result}\label{c.1-core-result}}

Let \((x_0, x_1, \dots, x_L)\) be a residual stream trajectory, and
define inter-layer steps:

\[
v_\ell = x_{\ell+1} - x_\ell \quad \text{for } \ell = 0, \dots, L-1
\]

Let \(v_\ell^{\text{ctrl}}\) be the step at layer \(\ell\) for the
control prompt, and define the shift-induced difference:

\[
\Delta v_\ell = v_\ell^{\text{shift}} - v_\ell^{\text{ctrl}}
\]

If there exists any layer \(\ell^*\) such that \(\Delta v_{\ell^*}\) is
\textbf{not colinear} with \(v_{\ell^*}^{\text{ctrl}}\), then the
concern-shifted trajectory exhibits strictly positive curvature at that
layer. That is, the path bends in a two-dimensional plane spanned by
\(\{v_{\ell^*}^{\text{ctrl}},\, \Delta v_{\ell^*}\}\).

Conversely, if \(\Delta v_\ell \parallel v_\ell^{\text{ctrl}}\) for all
\(\ell\), then the shifted trajectory lies in a single affine line and
all curvature vanishes.

\hypertarget{c.2-interpretation}{%
\subsubsection{C.2 Interpretation}\label{c.2-interpretation}}

This condition ensures that a geometric bend arises when the
shift-induced step differs in direction-not merely in magnitude-from the
control. It provides a discrete diagnostic: \textbf{directional
deviation implies curvature}.

\hypertarget{c.3-coordinate-considerations}{%
\subsubsection{C.3 Coordinate
Considerations}\label{c.3-coordinate-considerations}}

All measurements are taken in the residual space \(\mathbb{R}^d\), using
the semantic (pullback) metric \(G = U^\top U\) derived from the model's
unembedding matrix \(U\).

The curvature definition used in this study is invariant to
\textbf{orthogonal changes of basis} in \(\mathbb{R}^d\)-that is,
coordinate rotations do not affect the result. While such
transformations are not typical within transformer operations, this
mathematical invariance ensures that curvature is a property of the
trajectory's shape, not of any arbitrary axis labelling.

This result confirms that discrete, layerwise divergences can induce
true geometric bending-so long as they are directionally expressive. It
supports the use of \(\kappa_i\) as a valid and semantically aligned
curvature estimate.

\end{document}